\documentclass[sn-mathphys,Numbered]{sn-jnl}
\usepackage[utf8]{inputenc} 
\usepackage{mdframed}
\usepackage{hyperref}
\hypersetup{
    colorlinks=true,
    linkcolor=blue,
    filecolor=magenta,      
    urlcolor=cyan,
    pdftitle={AI4Extremes Review},
    pdfpagemode=FullScreen,
    }
\usepackage{comment}
\usepackage{graphicx}%
\usepackage{multirow}%
\usepackage{amsmath,amssymb,amsfonts}%
\usepackage{amsthm}%
\usepackage{mathrsfs}%
\usepackage[title]{appendix}%
\usepackage{xcolor}%
\usepackage{textcomp}%
\usepackage{manyfoot}%
\usepackage{booktabs}%
\usepackage{algorithm}%
\usepackage{algorithmicx}%
\usepackage{algpseudocode}%
\usepackage{listings}%
\usepackage{longtable}
\usepackage{multirow}

\usepackage{colortbl} 

\begin{document}

\title[AI for Extreme Events]{AI for Extreme Event Modeling and Understanding: Methodologies and Challenges}

\author*[1]{Gustau Camps-Valls}\email{gustau.camps@uv.es}
\author[1]{Miguel-\'Angel Fern\'andez-Torres}
\author[1]{Kai-Hendrik Cohrs}
\author[2]{Adrian H\"ohl}
\author[3]{Andrea Castelletti}
\author[16,4]{Aytac Pacal}
\author[5]{Claire Robin}
\author[13,14,15]{Francesco Martinuzzi}
\author[7,8]{Ioannis Papoutsis}
\author[7,8,1]{Ioannis Prapas}
\author[9]{Jorge P\'erez-Aracil}
\author[4,16]{Katja Weigel}
\author[1]{Maria Gonzalez-Calabuig}
\author[5]{Markus Reichstein}
\author[10]{Martin Rabel}
\author[3]{Matteo Giuliani}
\author[6]{Miguel Mahecha}
\author[10]{Oana-Iuliana Popescu} 
\author[1]{Oscar J. Pellicer-Valero} 
\author[11]{Said Ouala$^{\text{11}}$} 
\author[9]{Sancho Salcedo-Sanz}
\author[6]{Sebastian Sippel}
\author[7,8,1]{Spyros Kondylatos}
\author[12]{Tamara Happ\'e}
\author[1]{Tristan Williams}

\affil*[1]{\orgdiv{Image Processing Laboratory}, \orgname{Universitat de Val\`encia}, 
\country{Spain}}
\affil[2]{Chair of Data Science in Earth Observation, Technical University of Munich, Munich, Germany}
\affil[3]{\orgdiv{Department of Electronics, Information, and Bioengineering}, \orgname{Politecnico di Milano}, 
\country{Italy}}
\affil[4]{University of Bremen, Institute of Environmental Physics (IUP), Bremen, Germany}
\affil[5]{Max Planck Institute of Biogeochemistry, Jena, Germany}
\affil[6]{Leipzig University, Leipzig, Germany}
\affil[7]{National Technical University of Athens, Greece}
\affil[8]{National Observatory of Athens, Athens, Greece}
\affil[9]{Department of Signal Processing and Communications, Universidad de Alcal\'a, 
Madrid, Spain}
\affil[10]{German Aerospace Center (DLR), Jena, Germany}
\affil[11]{IMT Atlantique, Brest, France}
\affil[12]{Institute for Environmental Studies, VU Amsterdam, The Netherlands}
\affil[13]{Center for Scalable Data Analytics and Artificial Intelligence (ScaDS.AI), Leipzig, Germany}
\affil[14]{Institute for Earth System Science \& Remote Sensing, Leipzig University, Leipzig, Germany}
\affil[15]{Remote Sensing Centre for Earth System Research (RSC4Earth), Leipzig University, Leipzig, Germany}
\affil[16]{Deutsches Zentrum für Luft- und Raumfahrt (DLR), Institut für Physik der Atmosphäre, Oberpfaffenhofen, Germany} 

\abstract{
In recent years, artificial intelligence (AI) has deeply impacted various fields, including Earth system sciences. Here, AI improved weather forecasting, model emulation, parameter estimation, and the prediction of extreme events. However, the latter comes with specific challenges, such as developing accurate predictors from noisy, heterogeneous and limited annotated data. 
This paper reviews how AI is being used to analyze extreme events (like floods, droughts, wildfires and heatwaves), highlighting the importance of creating accurate, transparent, and reliable AI models. We discuss the hurdles of dealing with limited data, integrating information in real-time, deploying models, and making them understandable, all crucial for gaining the trust of stakeholders and meeting regulatory needs. We provide an overview of how AI can help identify and explain extreme events more effectively, improving disaster response and communication. We emphasize the need for collaboration across different fields to create AI solutions that are practical, understandable, and trustworthy for analyzing and predicting extreme events. Such collaborative efforts aim to enhance disaster readiness and disaster risk reduction.}


\maketitle

\tableofcontents

\section{Introduction} 
\label{sec:introduction}

The frequency, intensity, and duration of weather and climate extremes have increased in recent years, posing unprecedented challenges to societal stability, economic security, biodiversity loss, and ecological integrity \cite{seneviratne2021weather}. These events --ranging from severe storms and floods to droughts and heatwaves-- exert profound impacts on human livelihoods and the natural environment, often with long-lasting and sometimes irreversible consequences. Modeling, characterizing, and understanding extreme events is key to advancing mitigation and adaptation strategies.

In this context, Artificial Intelligence (AI) has emerged as a transformative tool for the detection \cite{aidetoolbox}, forecasting \cite{Lam2023,ferchichi2022}, analysis of extreme events and generation of worst-case events~\cite{ragone2021rare}, and promises advances in attribution studies, explanation, and communication of risk \cite{Yokota2004}. 
The capabilities of machine learning (ML), and deep learning (DL) in particular, in combination with computer vision techniques, are advancing the detection and localization of events by exploiting climate data, such as reanalysis and observations \cite{salcedosanz2024analysis}. 
Modern techniques for quantifying uncertainty \cite{gawlikowski2023survey} are a necessary step to making progress in climate change risk assessment \cite{harrington2021quantifying}. 
Using ensembles in combination with AI models has progressed the field of attribution of extremes \cite{stott2016attribution}, identification of patterns, trends \cite{madakumbura2021anthropogenic} and climate analogs \cite{chattopadhyay_analog_2020}.  
Yet, AI does not only excel at prediction but can also explain processes (e.g., via explainable AI \cite{FindingtherightXAIMethodBommer2024} and causal inference \cite{Hannart2016CausalCT}), which is essential for decision-making and effective mitigation strategies. 
Recent advances in large language models (LLMs) that retrieve information from heterogeneous text sources allow for effective integration of methods for communication tasks and human-machine interaction in extreme event and situation analysis \cite{vaghefi2023chatclimate}.

\begin{figure}[t!]
    \centering
    \includegraphics[width=12cm]{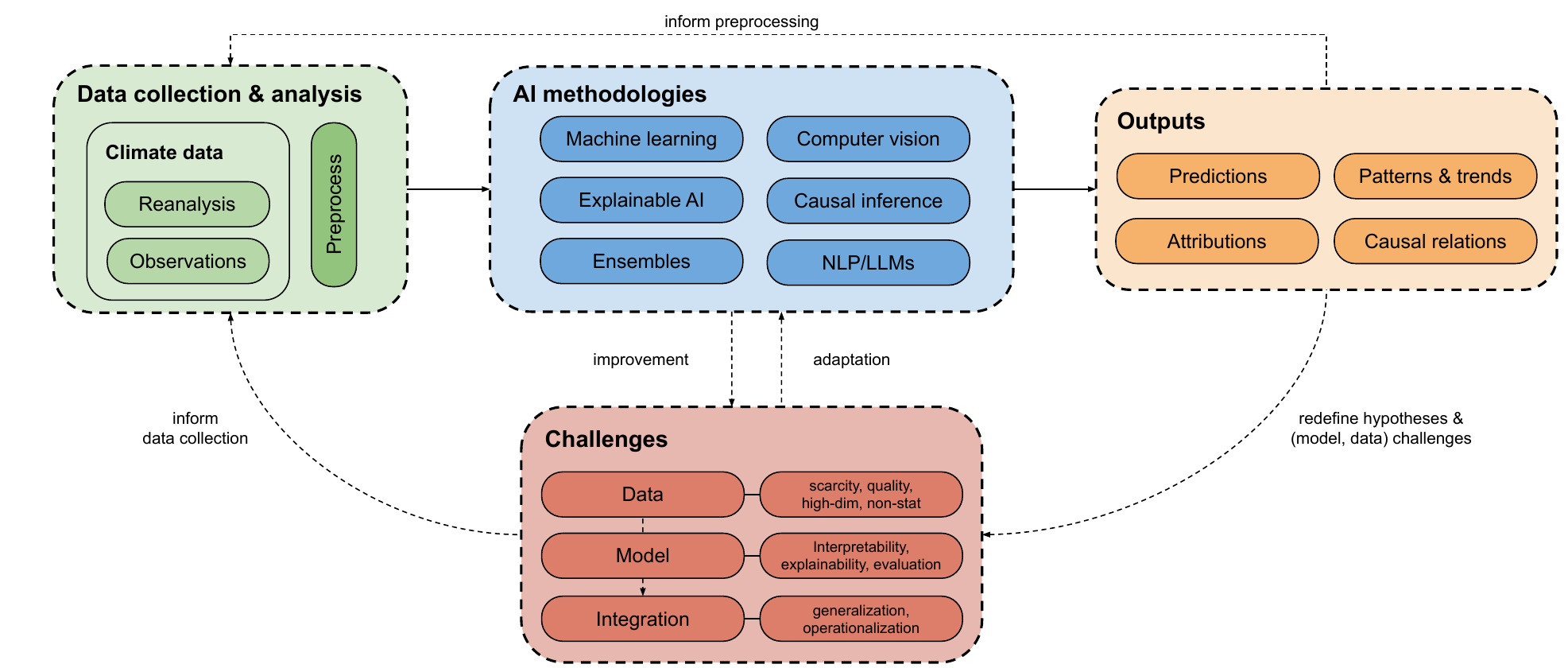}
    \caption{{\bf 
    A general AI-driven extreme event analysis pipeline.} Different components in modeling and understanding extreme events using AI methodologies are interconnected, highlighting the flow from data collection and analysis to actionable insights/outputs and the challenges encountered. Note the feedback loops where AI does not only produce some relevant outputs and products from data (predictions, patterns and trends, climate attributions, and causal relations) but also may help suggest areas for improvement and adaptation in methodologies to overcome identified challenges, redefine the hypotheses and challenges themselves, as well as inform data collection and preprocessing.
    }
    \label{fig:pipeline}
\end{figure}
Here, w review the role of AI for extreme event analysis, its challenges and opportunities for extreme event analysis. The general pipeline of AI-driven extreme event analysis (cf. Fig.~\ref{fig:pipeline}) encapsulates the entire workflow from data collection and preprocessing to the generation of outputs such as predictions, patterns, trends, climate attributions, and causal relations. The framework also highlights the iterative nature of AI processes, where outputs serve their direct purpose and also serve to inform and improve data collection, preprocessing, and hypothesis formulation. 
Nevertheless, developing AI models that are accurate, operational, explainable, and trustworthy still remains a significant challenge. Issues such as data scarcity, the need for high-dimensional data processing, and the interpretation and explanation of AI models are critical challenges that must be addressed. Moreover, communicating risk effectively and ethically, and integrating AI tools with existing systems to ensure operational viability and scientific robustness are nontrivial tasks requiring ongoing research and development.

We provide a comprehensive overview of the AI-driven pipeline for analyzing extreme events, including modeling, detection, forecasting, and communication (Section \ref{sec:survey}), and discuss key challenges (Section \ref{sec:challenges}). We present case studies on AI applications for droughts, heatwaves, wildfires, and floods (Section \ref{sec:cases}), and conclude by identifying major challenges and future research opportunities in AI for extreme events (Section \ref{sec:conclusions}).

\section{Review of AI methods}
\label{sec:survey}

This section reviews the main methods in all aspects of extreme event analysis (cf. Table~\ref{tab:challenges}): data, modeling, understanding and trustworthiness, and the last mile on communication and deployment. 

\begin{table}[t!]
\caption{A comprehensive categorization of extreme events analysis in four main blocks (data, modeling, understanding and trustworthiness, and the last mile on communication and deployment) regarding their task, standard and current AI methods.}
    \label{tab:challenges}
\centerline{\includegraphics[trim={0 10cm 0 0 0},clip,width=12cm]{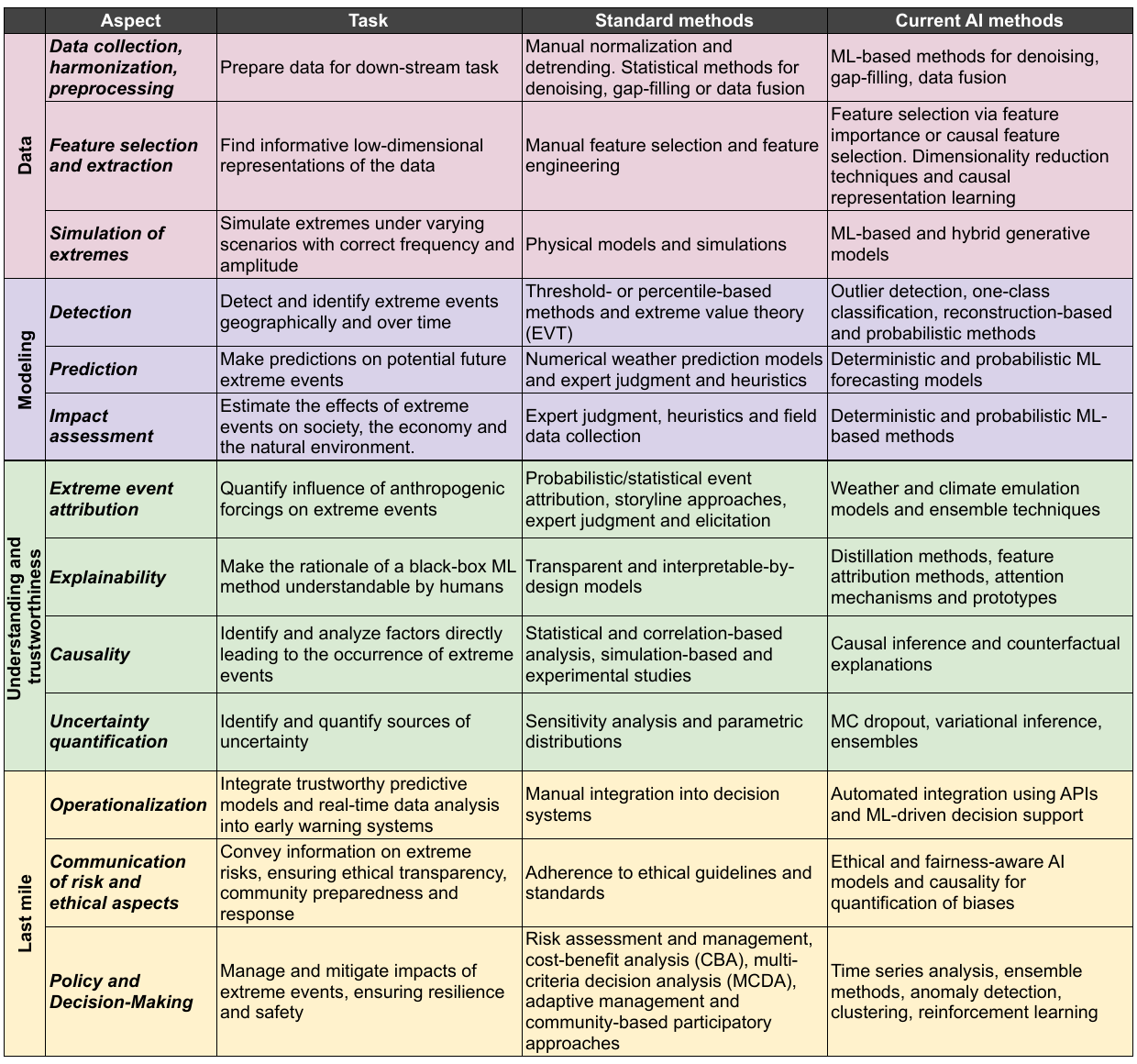}}
\end{table}

\subsection{Extreme Event Modeling}

AI methodologies for extreme event modeling can be categorized into detection \cite{ruff2021unifying}, 
prediction \cite{han2019review} and impact assessment \cite{zennaro2021exploring} tasks. 
Thanks to the emergence and success of DL, all of these tasks can be tackled by designing data-driven models that exploit spatio-temporal and multisource Earth data characteristics, from climate variables 
to {\em in situ} measurements and satellite remote sensing images (see Figure \ref{fig:components} [top row]).

\subsubsection{Detection}

Detecting and identifying extreme events geographically over time to distinguish them from normal conditions is fundamental for discovering underlying patterns and correlations between climate variables. This allows for a better understanding of their generating processes and mechanisms and facilitates their anticipation. 

Classical statistical methods, such as threshold or percentile-based indices, have been widely applied to detect extreme events (see Table~\ref{tab:challenges}). However, these often fail to identify the same events as expert or impact-based detection methods \cite{Jones_2023}. AI methodologies can help reconcile expert knowledge with data-driven approaches as they capture regularities and nuanced relations in large volumes of observational data. 
Canonical ML treats the detection problem as a one-class problem or an outlier detection problem. Many methods are thus applied \cite{flach2017multivariate}, and available in software packages \cite{aidetoolbox}. Recent advances include deep learning for segmentation and detection of tropical cyclones (TCs) and atmospheric rivers (ARs) in high-resolution climate model output \cite{karaismailoglu2021climatenet} and semi-supervised localization of extremes \cite{racah2017extremeweather}. Alternatively, reconstruction-based models (e.g., with autoencoders) are optimized to accurately reconstruct normal data instances, and thus extremes are associated with large reconstruction errors \cite{GuancheGarca2018}. 
Finally, probabilistic approaches try to identify extremes by estimating the data probability density function (PDF) or specific quantiles thereof. 
Standard Extreme value theory (EVT) approaches, however, struggle with short time-series data, nonlinearities, and nonstationary processes. Alternative probabilistic (ML) models have relied on surrogate meteorological and hydrological seasonal re-forecasts \cite{Klehmet2024}, Gaussian processes and nonlinear dependence measures \cite{Johnson20sakame} and multivariate Gaussianization \cite{Johnson20rbig4eo}.

\subsubsection{Prediction}

Designing predictive systems that accurately model extreme events is essential to anticipating the effects of future extreme events and providing critical information for decision-makers to prevent damages. Spatial and temporal predictions aim to provide a quantitative estimate of the future value of the Earth's state (Table~\ref{tab:challenges}). 

Many ML algorithms have been proposed for deterministic extreme event prediction, but most are only applied to small regions and specific use cases, though. Prediction can be performed using climate variables alone \cite{vijverberg2022role} or in combination with satellite imagery \cite{kladny2024enhanced}. Here, for example, the vegetation response under extreme drought conditions is modeled as a video prediction task of the satellite imagery while conditioning on the climate variables. Another common approach is to directly estimate an indicator that defines the extreme event, e.g. flood risk maps \cite{bentivoglio2022deep} or drought indices \cite{belayneh2014long}. The estimated variables can cover different lead times depending on the characteristics of the extreme, spanning from short-term prediction to seasonal prediction \cite{kondylatos_wildfire_2022}. 
Recently, DL-based prediction techniques have gained popularity for their ability to process large data volumes, capture complex nonlinear relationships, and reduce manual feature engineering. These benefits have led to the creation of global models that generalize across various locations, as seen in flood \cite{nearing2024global} and wildfire \cite{zhang2021deep} predictions.

Probabilistic models, unlike deterministic ones, focus on predicting the probability distribution of future variable states. The importance of probabilistic forecasts for extreme heatwaves has been emphasized \cite{miloshevich2023probabilistic}. 
AI-based techniques, integrated within climate models, can enhance predictions, such as in drought prediction \cite{vo2023lstm} and extreme convective precipitations downscaling \cite{shi2020enabling}.

\subsubsection{Impact Assessment}

Estimating the effects of extreme events on society, the economy, and the environment is crucial for conveying potential future consequences to the public, policymakers, and across disciplines \cite{callaghan2021machine, zennaro2021exploring}. Impact assessment involves understanding how a system reacts to extreme event forcings. Unlike extreme event detection and prediction, the focus here is on impact-related outcomes, such as the number of injuries, households affected, or crop losses.

In recent years, there has been an increasing interest in predicting vegetation state using ML \cite{ferchichi2022}, as a way to estimate the impact of climatic extremes on the evolution of the vegetation state variable 
\cite{sutanto2019moving,salakpi2022forecasting}. 
Recent advances methods have used echo state networks \cite{martinuzzi2023learning}, 
ConvLSTM-based \cite{ahmad2023machine,kladny2024enhanced}, 
and transformers \cite{benson2023forecasting}
using high-resolution remote sensing and climate data. 

Another way to address impact assessment with ML is to analyze changes in the probability density function over time. This approach allows a quantification of the impact of different events, which can be used to improve our understanding of the drivers of vulnerability, \cite{zhang2021impact} to 
such as population displacement \cite{ronco2023exploring}. Alternatively, the impact of extreme events can be detected by analyzing the news coverage based on natural language processing (NLP) and, more recently, on large language models (LLMs) \cite{Sodoge2023}. 

\begin{figure}[t!]
    \centering
    \includegraphics[width=12cm]{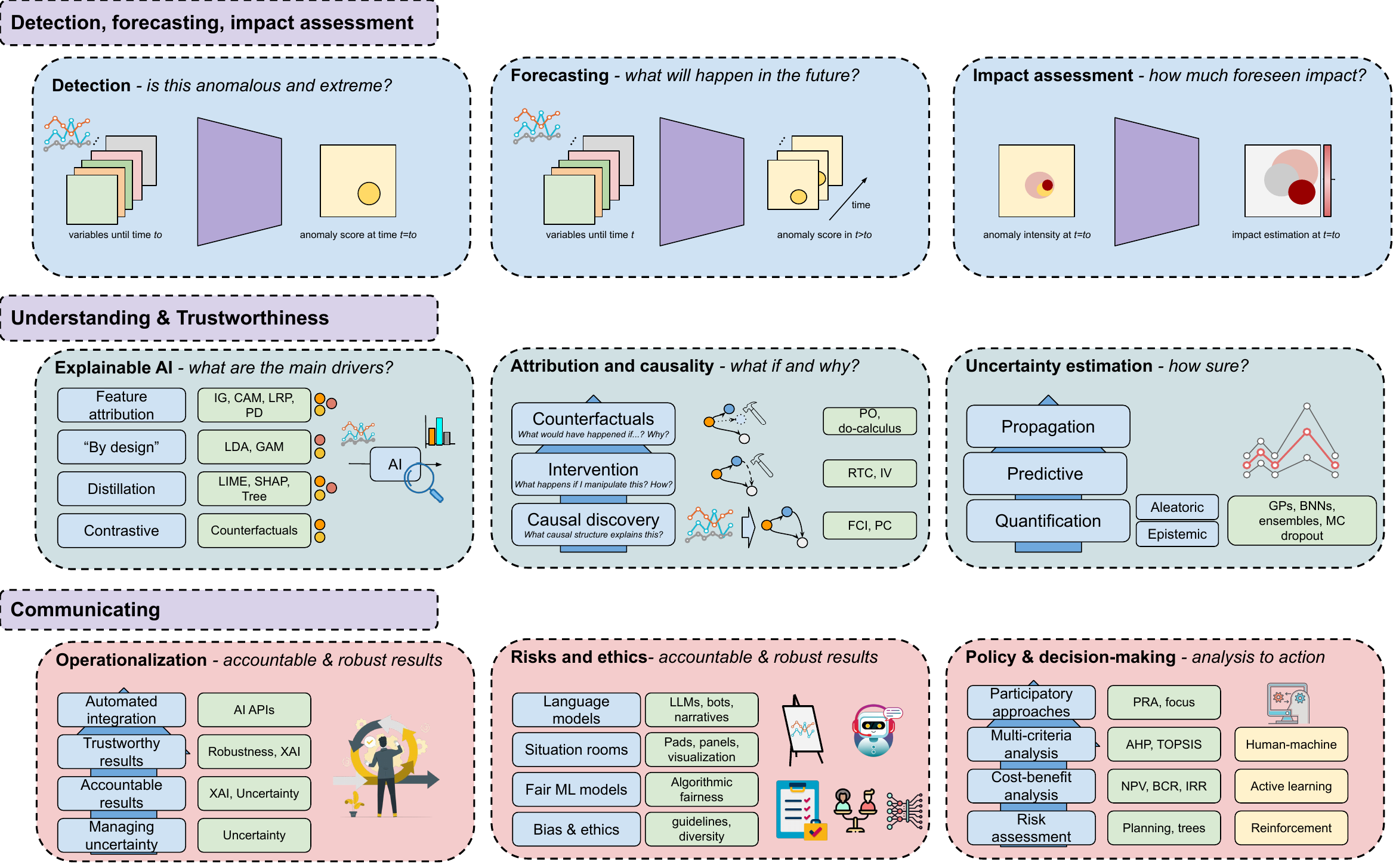}
    \caption{{\bf Components in an AI pipeline for extreme events.} 
    AI mainly exploits spatio-temporal Earth observation, reanalysis, and climate data to answer ``what'-questions (top row): detection of events, prediction, and impact assessment. 
    AI can also be used for understanding events and thus answer ``what if,'' ``why,'' and ``how sure'' questions (middle row) and makes use of explainable AI (XAI) to identify relevant drivers of events, causality to understanding the system, estimate causal effects and impacts, and imagine counterfactual scenarios for attribution and uncertainty estimation to quantify trust and robustness for decision-making. 
    Communicating extreme events and their impacts can benefit from statistical/machine learning (bottom row) by improving operationalization, ensuring fair and equitable narratives, and integrating language models in situation rooms for enhanced decision-making.}
    \label{fig:components}
\end{figure}

\subsection{Extreme events understanding and trustworthiness}

Yet, all previous approaches focus on the `what,' `when,' and `where' questions, not the `why,' `what if,' and `how certain' ones.
In environmental sciences, there's an increasing emphasis on the need for trustworthy machine learning \cite{bostromTrustTrustworthy}. This is especially important in extreme events, where critical high-stakes decisions need to be made impacting public safety, health, infrastructure, and resource allocation. Disciplines like explainable AI (xAI) and uncertainty quantification (UQ) offer methods to make AI more reliable and trustworthy (see Figure \ref{fig:components} [middle row]). These approaches not only help us interpret AI model predictions but also enhance our understanding of extreme events themselves. Techniques such as causal inference and extreme event attribution further enhance xAI and UQ by understanding the mechanisms behind these events, crucial for improving AI models and gaining trust in the decision-making process.

        \subsubsection{Extreme Event Attribution}

        Extreme event attribution (EEA) quantifies the influence of anthropogenic forcings (such as greenhouse gas emissions) on the likelihood of extreme climate events by using numerical simulations with General Circulation Models (GCM) to compare their probabilities under observed conditions (the factual world) and a hypothetical scenario without human emissions (the counterfactual world) \cite{naveau2020statistical} (cf. Table~\ref{tab:challenges}). 
        Two main viewpoints exist: probabilistic EEA employs quantitative statistical methods to estimate this likelihood (e.g., the EE is 600 times more likely due to human emissions), while storyline approaches simulate the evolution of the EE under different forcings to gather a process-based attribution statement (e.g., 50\% of the magnitude of an EE is explained by natural variability) \cite{otto2023attribution}. Ideally, both methods should be combined to provide a complete understanding.

        Ensembles of neural networks have been developed to emulate GCMs, creating surrogate models that operate significantly faster \cite{pasini2017attribution}. These models successfully predict the year based on the global annual temperature field under present conditions but struggle to perform accurately in hypothetical scenarios resembling pre-industrial conditions \cite{barnes2019viewing}.
        Despite the scarce literature on this topic, the impressive recent advances in weather and climate emulation with AI \cite{Lam2023} suggest that climate emulators will play a significant role shortly, allowing almost real-time EEA. 
        
        \subsubsection{Explainable AI} 
        \label{sec:XAI}
        Many models are transparent and interpretable-by-design, like linear models or decision trees, but they may not perform well on complex problems and are rarely employed in extreme events (cf. Table~\ref{tab:challenges}).
        Explainable AI (xAI) aims to unveil the decision-making process of AI models. 
        xAI also facilitates debugging, improving models, and gathering scientific insight by revealing the model functioning, learned relationships, and biases. 
        The most commonly used xAI approaches rely on model-agnostic {\em distillation} or {\em feature attribution} methods \cite{ghaffarianExplainableArtificial2023,Tuia24perspective}, cf. Fig.~\ref{fig:components}. Distillation methods, such as SHapley Additive exPlanations (SHAP) and Local interpretable model-agnostic explanations (LIME), create surrogate models and have been widely used in geosciences and climate sciences \cite{schlund2020constraining,ronco2023exploring}. 
        Feature attribution methods, like Partial Dependence Plot (PDP) or Gradient-based Class Activation Map (Grad-CAM), highlight important features by perturbing the inputs or using backpropagation \cite{srinivasan2020machine}.         
        Recent approaches explain DL models using 
        attention in drought prediction \cite{dikshitSolvingTransparency2022a} and
        prototypes for explaining event localization \cite{barnesThisLooks2022}. 
        xAI have been also used to evaluate climate predictions \cite{FindingtherightXAIMethodBommer2024, mamalakisInvestigatingFidelity2022}, and thus offering an agnostic, data-driven approach for model-data intercomparison.

        \subsubsection{Causal inference}
        
        Causal graphs encode intuitive information about robust system properties. This information is further required for deciding the identifiability of queries about interventional or counterfactual properties \citep{pearl2017causality,Peters2017elements}. 
       For extreme events, most widespread causal inference methods \cite{Runge2023,CampsValls23physcausaldiscovery} cannot be applied directly, as traditional assumptions of normality and mild outliers do not hold, and dependence may manifest only in the extremes (Table \ref{tab:challenges}). This has motivated recent works that deal with causal discovery for extreme event analysis, providing different frameworks for understanding the dependencies and causal structures in extreme values \cite{Gnecco2019CausalDI}, with applications on, e.g. rivers discharge \cite{Pasche2021CausalMO}. Another relevant line of work focuses on answering counterfactual questions for climate extreme events, which links to EEA \citep{Hannart2016CausalCT, Kiriliouk2019ClimateEE}.  
               
        \subsubsection{Uncertainty quantification} 
        
        Even equipped with explanatory and causal methods, it is still crucial to assess how confident are AI model decisions, as inaccurate warnings or decisions could impact safety and resources~\citep{gawlikowski2023survey,ghanem2017}. 
        Understanding the sources of uncertainty is important to inform and disentangle the inherent (aleatoric) uncertainty of the weather phenomenon from the lack of knowledge in the model (epistemic) uncertainty. One can reduce the latter with more data and additional assumptions, but not the former (cf. Table~\ref{tab:challenges}). 
        DL approaches with UQ have shown promise for extreme events recently~\cite{miloshevich2023probabilistic, xu2022quantifying}. 

    \subsection{The Last Mile: operationalization, communication, ethics and decision-making}

    The previous components (methods and techniques) in the AI pipeline need to be operationalized,     be robust, accountable and fair, 
    so they ultimately serve the purpose of evidence-based policy making (see Figure \ref{fig:components} [bottom row]). 
    
\subsubsection{Operationalization}

Operationalization strictly needs the previous layer in place (xAI, causality, UQ) for improved 
and accountable explanations of the predictions in early warning systems (EWSs), 
and for enhanced disaster risk management.  
Calibration of ML models \cite{bella2010calibration}, especially when coupled with uncertainty estimation, aligns predicted probabilities to reflect the real-world likelihood of extreme outcomes, enhancing reliability and interpretability essential for informed decision-making. Poor calibration can lead to misguided decisions, such as overestimating or underestimating the likelihood of critical events.

\subsubsection{Communication of risk and ethical aspects} 
History shows that even actionable forecasts can fail if not communicated properly. For instance, despite predictions of the Mediterranean storm Daniel's landfall four days in advance, the lack of effective communication contributed to the tragic outcome in Libya, with severe casualties and displacement \cite{corps2023libya}. 
Even in the more developed country of Germany, the devastating effects of floods 
were related to ineffective warnings \cite{dwd2021weather}. 
This highlights the critical need for robust EWSs that predict events and effectively communicate risks to ensure community preparedness and response (Table~\ref{tab:challenges}).

The challenge of false alarms in the context of EWS 
is a significant challenge, as they can lead to ``warning fatigue'', where the public becomes desensitized to alerts and may ignore crucial warnings during actual emergencies 
\cite{tojcic2021performance, yore2020early,reichstein2023ews}. Addressing this issue requires improving the accuracy and trust in  predictive models, 
refining communication strategies, and engaging the community \cite{tamamadin2020automation}.

\subsubsection{Ethics in AI models and data}

The governance of AI ethics calls for systems to be respectful of human dignity, ensure security, and support democratic values \cite{RNY004}. The deployment of AI to help manage extremes involves several fundamental principles: ensuring fairness, maintaining privacy, and achieving transparency \cite{RNY003,kochupillai2022earth}. In this context, the rise of Large Language Models (LLMs) heightens ethical risks like outdated or inaccurate data, bias, errors, and misinformation, often obscured by their vast training on web data. 
Additionally, current AI models (like generative AI and LLMs) often draw on large datasets, so there is a risk of strong biases if these datasets are unrepresentative \cite{Tuia24perspective}. 
Spatial sampling and analysis are vital to collecting geographically and environmentally representative, fair, and unbiased data. 
Global initiatives like the Common Alerting Protocol have standardized warning data, but the success of these systems depends on their inclusivity and ability to adapt to the diverse needs of affected communities, who should participate in developing EWSs 
\cite{macherera2016review,reichstein2023ews}. AI enhances these systems by enabling the rapid dissemination of personalized alerts tailored to specific locations and individual risk factors like proximity to flood plains or wildfire zones, ensuring warnings are understandable and relevant to all.

Achieving the inclusion of affected communities, especially in the Global South, still 
remains a significant challenge, as centralized, 'one-size-fits-all' models are often simpler and cheaper to implement but less effective at addressing local nuances. Designing AI systems, building on large-language models fine-tuned to local communities with users in the loop, offers excellent opportunities to overcome ``one-size-fits-all'' while being efficient. 

\subsubsection{Policy and decision-making} 

Even if AI-assisted, human operators are in charge of implementing the final decision. 
This implies supporting end-users in making the most out of the generated information 
\cite{Yokota2004}. This operational value is often problem-specific and depends upon the system’s dominant dynamics and the socio-economic sectors considered. 
Quantifying this value, however, might not be straightforward as a more accurate prediction does not necessarily imply a better decision by the end users \cite{Ramos2013}. 
When multiple forecasts from different systems are available, users should address a number of problems where AI can be helpful, including the selection of the forecast product, the lead time, the variable aggregation, the bias correction, and how to cope with the forecast uncertainty.

Integrating the pipeline in Fig. \ref{fig:pipeline} with impact models that leverage Reinforcement Learning algorithms to simulate optimal decision-making can help quantify how AI-enhanced information about extreme events translates into better decisions \cite{Giuliani2015ISA}. The co-design of impact models via participatory processes, including end users in the loop, 
further strengthens the overall model-based investigation by better capturing end users' requirements, expectations and concerns. 
End users should understand both the benefits of AI-enhanced information for improved decision-making and the risks of misinformation leading to poor choices.

\section{Data, model and integration challenges}
\label{sec:challenges}

Extreme event analysis faces many important challenges related to the data and model characteristics, but also to the integration of AI in decision pipelines (see Table~\ref{tab:challenges_risks_future}.

\begin{table}[t!]
\caption{Challenges and risks of using AI for extreme event analysis, and needed advances in AI to tackle them. The same blocks before related to data, model and integration challenges are studied. }
    \label{tab:challenges_risks_future}
\centerline{\includegraphics[width=12cm]{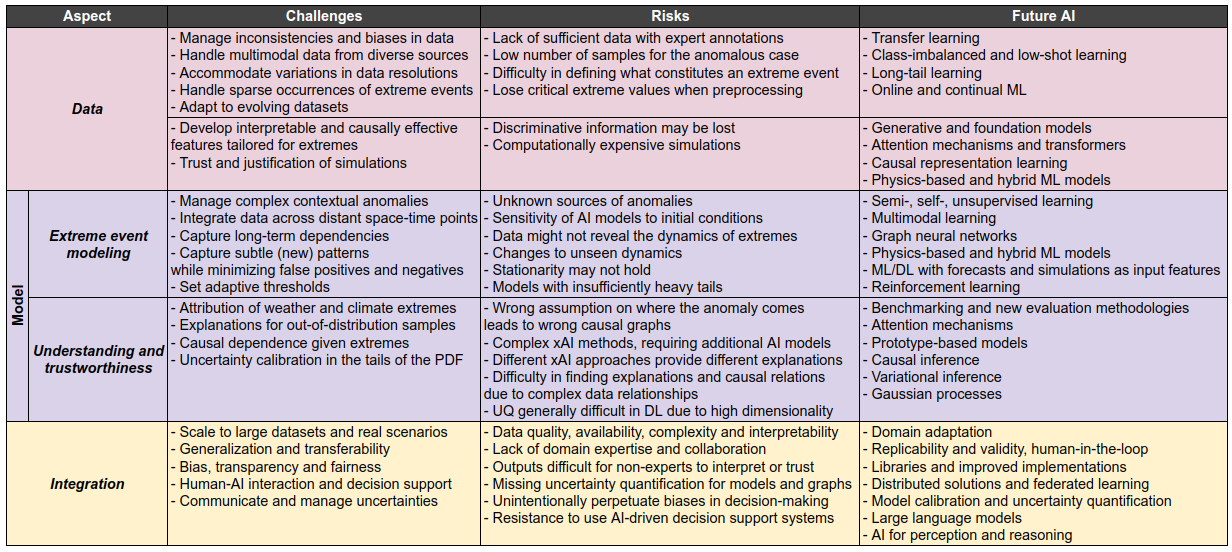}}
\end{table}

\subsection{Data challenges}

A major challenge is the lack of sufficient data with expert annotations, which are essential for training and evaluating AI models. Given their rarity, extreme events can be overlooked during the data preprocessing steps to eliminate noise, gaps, biases, and inconsistencies \cite{camps2021deep,Reichstein19nat}. Additionally, AI struggles with integrating and extracting relevant information across various data sources and scales, which can complicate feature extraction and selection \cite{Tuia24perspective}. Future AI development needs to focus on deriving robust features (or representations) that effectively capture the distinct characteristics of extreme events (cf. Table~\ref{tab:challenges_risks_future}). 

AI models are increasingly used to enhance parameterizations of subgrid-scale processes in Earth system models, addressing gaps in traditional methods. However, a key challenge is their numerical instability over extended time horizons, which may produce unrealistic scenarios when simulating extremes due to insufficient training data \cite{iles2020benefits}. 
The quality of observational data also challenges AI methods used for data assimilation. Recently, hybrid ML models that combine domain-driven with data-driven models approaches promise more robust and trustworthy AI models \cite{cheng2023machine,Reichstein19nat}. 
Moreover, AI-based methods have improved error characterization, enhancing uncertainty quantification \cite{Haynes2023}. Generative models are also used to sample ensemble members more effectively, offering better representations of the system's state \cite{Li2022} (cf. Table~\ref{tab:challenges_risks_future}).

\subsection{Model challenges}

The lack of a clear statistical definition of extreme events and the mechanisms responsible for their occurrence hamper model development and adoption. 
For detection, extremes usually constitute not pointwise but complex contextual, group, or conditional anomalies, whose source (changing process, parents, distribution) is often unknown \cite{Tuia24perspective}. This results in challenges, such as capturing subtle (new) patterns, setting adaptive thresholds, or integrating data across distant points in space and time (cf. Table~\ref{tab:challenges_risks_future}).
For prediction and impact assessment, AI models are sensitive to initial conditions and may not capture long-term dependencies \cite{camps2021deep}. Moreover, data might not reveal the dynamics of extremes, there might be changes to unseen dynamics \cite{flach2017multivariate}, and stationarity may not hold in general. 
Using hybrid models that combine data, domain knowledge and models could allow insights into the mechanisms that trigger extreme events \cite{Reichstein19nat,farazmand2019extreme,cheng2023machine}.

The complexity of extremes also makes 
attribution, causal discovery, and explainability particularly challenging. 
xAI can only reveal correlations the model learned and has no information about the causal structure. This could lead xAI to exacerbate   
model biases or spurious correlations. Indeed, different xAI methods can produce very different explanations, whose suitability for different models should be quantitatively evaluated \cite{FindingtherightXAIMethodBommer2024}.  
Causality is not error-free, as a wrong assumption about where the extreme comes from could lead to wrong causal graphs, conclusions and decisions \cite{pearl2017causality}. 
Finally, challenges in UQ include differentiating and quantifying aleatoric and epistemic uncertainties, which is complicated by the models' overparameterization and their lack of robust probabilistic foundations (cf. Table~\ref{tab:challenges_risks_future}). 

\subsection{Integration challenges}

ML models are typically trained on high-quality, well-curated datasets, such as Copernicus ERA 5 reanalysis or cloud-free satellite imagery, which often do not reflect the error-prone meteorological forecasts and cloudy conditions encountered in real-world situations (Table~\ref{tab:challenges_risks_future}). 
Applying domain adaptation strategies or leveraging invariant features could align model performance from training phases to operational conditions.
Besides, leveraging proprietary and trusted geospatial data from operational stakeholders, such as detailed forest fuel maps and elevation models, allows for fine-tuning these models to enhance detection and forecasting accuracy and enable finer spatial and temporal resolution in output products.

\section{Case Studies}
\label{sec:cases}

\begin{figure}[h!]
    \centerline{\includegraphics[width=12cm]{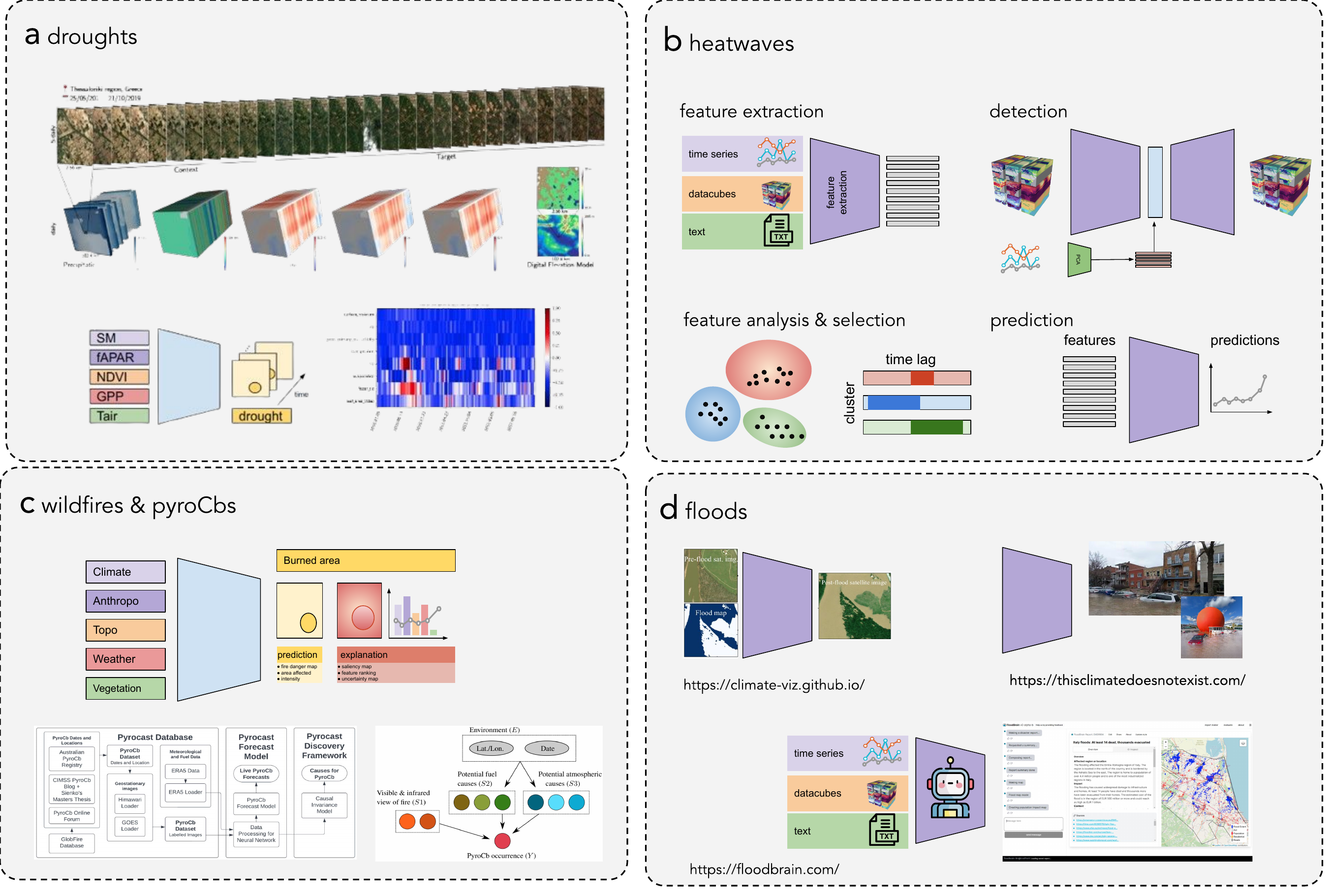}}
     \caption{\scriptsize {\bf Summary of case studies using AI to manage extreme events.} Four use cases (drought, heatwaves, wildfires, and floods) are showcased where AI enables detection, forecasting, impact assessment, explanation, understanding, and communication of risk, providing a comprehensive solution for disaster management. 
     {\bf (a) Droughts.} Top: AI leverages multimodal data to predict Earth's surface dynamics, enhancing forecasts for crop yields, forest health, and drought impacts. 
     Bottom: XAI techniques, like ``neuron integrated gradients,'' elucidate the key factors driving severe drought conditions, highlighting variable interactions over time.
     {\bf (b) Heatwaves.} Top: Variables of interest are extracted from heterogeneous data sources (images, time series, text) and potentially aggregated over space and/or time. 
     Bottom Left: Relevant features can be extracted from the data using e.g., clustering techniques. 
     Right: Prediction of heatwaves can be done in combination with dimensionality reduction tools or directly from the selected features.
     {\bf (c) Wildfires.} Top: AI enhances understanding and prediction of wildfire dynamics, particularly for mega-fires intensified by global warming, by analyzing extensive datasets and differentiating fire types with XAI. 
     Bottom: AI combined with causal inference aims to better detect and understand pyrocumulonimbus clouds, intense storm systems generated by large wildfires that complicate fire behavior prediction. 
     {\bf (d) Floods.} AI transforms flood risk communication by using realistic 3D visualizations and animations to depict rising water levels' impact on communities and infrastructure, making the information more relatable ({\tt thisclimatedoesnotexist.com}). AI-driven platforms analyze vast amounts of data from weather forecasts, river levels, and historical flood patterns to predict future events accurately, integrating this information with digital maps and urban models to identify high-risk areas ({\tt climate-viz.github.io.com}). This approach enhances flood risk management by allowing for targeted, personalized communication, enabling residents to receive specific alerts and visualize potential impacts on their homes. AI also supports the generation of detailed flood reports from various sources, enhancing preparedness and mitigation efforts ({\tt floodbrain.com}). }
    \label{fig:casestudies}
\end{figure}

We showcase four case studies on drought, heatwaves, wildfires, and floods, each of them covering the different aspects of detection, forecasting, impact, explainability, attribution, and communication of risk (Fig.~\ref{fig:casestudies}).

\subsection{Droughts: detection, forecasting and explainability} 

Droughts are among the costliest natural hazards, having destructive effects on the ecological environment, agricultural production, and socioeconomic conditions. They have been traditionally approached with heuristic thresholds and simple parametric models on related variables (e.g., soil moisture, vegetation indices, precipitation, or temperature, etc.). Being drought monitoring historically purely based on station-based measurements, Earth observation technology (e.g., Sentinel and Landsat series) and remote sensing allowed for estimating drought-related variables over larger spatial and temporal scales \cite{west2019remote}. However, drought definitions presented in the literature (meteorological, hydrological, agricultural, operational, or socioeconomic) are usually subjective and limited \cite{zargar2011review}. These types are not independent but refer to complementary physical, chemical, and biological processes of the Earth system droughts involve. Although empirical multivariate drought indicators have been proposed to account for these dependencies, the complexity of the combination of drought processes makes their detection, prediction, and characterization (severity, duration, start-end, etc.) a burden to researchers and policymakers, which can benefit from data-driven AI approaches. 

Traditional ML approaches have been successfully applied to drought prediction, from support vector machines, decision trees, and random forests to more advanced multivariate density estimation approaches \cite{jehanzaib2021comprehensive,2024mengxue,Johnson20rbig4eo}. However, DL algorithms are required for efficiently modeling spatial and temporal correlations in data. 
To this end, neural networks have been used for drought monitoring in a supervised way, mainly using multilayer, convolutional, and/or recurrent neural networks \cite{aidetoolbox}. 
Drought detection is an unsupervised problem due to its spatiotemporal variability and the shifting definition of drought under climate change. Thus, unsupervised and semi-supervised methods are recommended for practical applications \cite{ruff2021unifying}. 

To overcome some of the challenges before, recent works have applied AI methods to process multimodal data (satellite imagery at different resolutions, mesoscale climate variables, and static features). Drought detection has been addressed at the European level employing domain knowledge-driven variational autoencoders (VAEs) \cite{2024mengxue}, which combine traditional drought indicators with climate data.
Forecasting the dynamics of the Earth's surface with AI may help predict crop yield, forest health, and the effects of drought events \cite{benson2023forecasting} (Fig.~\ref{fig:casestudies}a, top). 
Similar DL drought detectors can be explained with xAI, identifying the most salient regions in time and space and attributing their main drivers \cite{dikshit2021interpretable} (Fig.~\ref{fig:casestudies}a, bottom). 

\subsection{Heatwaves: prediction, attribution and understanding}

In the context of climate change, heat extremes are becoming more frequent and intense \cite{barriopedro_heat_2023}, 
but still regional trends are uncertain \cite{teng2022warming}. For example, a notable warming hotspot in western Europe was identified with heat extremes increasing faster than predicted by the current state-of-the-art climate models \cite{vautard_heat_2023}. 
Understanding the physical drivers of heat extremes is crucial for accurate seasonal prediction 
and projection of these events. Several drivers of heat extremes remain uncertain though, specifically long-term changes in atmospheric circulation as well as land-atmosphere interactions, that may amplify or enhance specific events \cite{barriopedro_heat_2023}. 

AI methods can accurately detect and predict heat extremes on different spatial and temporal scales, some lead times and aggregations 
using deep learning  \cite{chattopadhyay_analog_2020, jacques2022deep, fister_accurate_2023}, causal-informed ridge regression \cite{vijverberg2022role}, or hybrid models \cite{van_straaten_CorrectingSubseasonal_2023}, among others. These approaches often require dimensionality reduction tools to select predictors from high-dimensional climate data (Fig.~\ref{fig:casestudies}c).  
Expressive lower-dimensional feature representations have been extracted from high-dimensional data using VAEs, which can be used to improve the prediction of heat extremes  \cite{miloshevich2023probabilistic}. 
New methods, such as the EarthFormer, explore transformer networks specifically for the prediction of temperature anomalies, where encoder/decoder structures are combined with spatial attention layers \cite{gao2023earthformer}. Next to predicting heatwaves, physical drivers of heatwaves can be inferred from the ML models as well. For example, XAI tools can be applied to discover physical drivers of heat extremes, e.g. \cite{van_straaten_CorrectingSubseasonal_2023}. Apart from XAI, causal inference methods have been used to understand the drivers of extreme temperatures \cite{vijverberg2022role}.

Attribution studies for temperature and precipitation have increasingly used AI techniques \cite{barnes2019viewing}. When it comes to heatwaves the identification of circulation-induced vs. thermodynamical changes becomes crucial. Several statistical and ML methods are currently in use for this purpose \cite{singh_circulation_2023}. 
The next vital step for attribution studies is to use such ML methods to understand the contribution of dynamical changes to the observed trends. This will lighten the discrepancies between circulation trends in climate model observations and reanalysis \cite{vautard_heat_2023}. Finally, recent development shows that ML techniques can be successfully applied for rare event modeling, sampling, and prediction \cite{miloshevich2023probabilistic}. 
    
\subsection{Wildfires: predict, explain and understand} 

Most wildfire damage stems from a few extreme events. Predicting these events is vital for effective fire management and ecosystem conservation. Climate change is expected to increase the frequency, size, and severity of extreme wildfires, exacerbating fire weather conditions \cite{jones2022global}. Traditional firefighting approaches are increasingly inadequate, with many extreme fires burning until naturally extinguished \cite{el_garroussi_europe_2024}. The complexity of nonlinear interactions among fire drivers across various scales hinders the predictability of fire behavior. Therefore, enhancing models to better understand and predict conditions leading to large, uncontrollable fires is crucial.

DL methods can associate weather predictions, satellite observations, and burned area datasets to model wildfires achieving better predictability than conventional methods. DL has been used successfully to forecast wildfire danger \cite{kondylatos_wildfire_2022} 
and for susceptibility mapping \cite{zhang_forest_2019}. In the short term, wildfires are driven by the daily weather variations but also by the cumulative effects of vegetation and drought. On sub-seasonal to seasonal scales, where weather forecasts are less reliable, wildfires are modulated by Earth system large-scale processes, such as teleconnections. Early work has shown that DL models can leverage information from teleconnections to improve long-term wildfire forecasting \cite{li2023attentionfire_v1}. Apart from predicting wildfires, it is very important to understand the reasons behind the predictions to support decision-making and fire management. In this context, xAI can aid in identifying what drives wildfires, supporting, for example, the differentiation and management of various fire types, such as wind-driven and drought-driven fires \cite{kondylatos_wildfire_2022}. 

Under global warming \cite{el_garroussi_europe_2024}, in the coming years, we expect an increase in the frequency of unstable atmospheric conditions that can cause Pyrocumulonimbus (pyroCbs), i.e., storm clouds that generate their weather fronts and can make wildfire behavior unpredictable \cite{fromm_understanding_2022}. Despite the risk posed by pyroCbs, the conditions leading to their occurrence and evolution are still poorly understood, and their causal mechanisms are uncertain. AI, in combination with causal inference, can advance the detection, forecasting, and understanding of the drivers of pyroCb events \cite{salas2022identifying}.

\subsection{Floods: from detection to communication of risk} 

Studying floods is crucial as they are the most frequent and costly natural disasters, affecting millions annually and causing over \$40 billion in damages worldwide each year \cite{Jonkman2005Global}. Developing novel methods for flood detection enhances EWSs, reducing fatalities by up to 40\%, while effective risk communication ensures better preparedness and response, potentially saving thousands of lives. 

In July 2021, intense rainfall in Germany, Belgium, France, and the UK triggered severe flash floods, particularly devastating the Ahr River region in Germany, claiming nearly $200$ lives \cite{cornwall2021europe,mohr2023multi}. Alerts were issued by the European Flood Awareness System (EFAS) and disseminated through an international warning system. However, the floods exposed significant flaws in these systems (Fig.~\ref{fig:casestudies}d). Damage to river monitoring infrastructure impeded data accuracy, and although meteorological forecasts predicted the rainfall two days in advance, flood predictions for smaller basins lacked precision and failed to consider debris flow and morphodynamics, leading to underestimations of flood severity.

AI offers promising avenues for enhancing flood management systems. Advanced global meteorological forecasting models powered by AI can rapidly process large data ensembles, providing more accurate probabilistic estimates even during extreme weather events. Furthermore, AI techniques such as ML-accelerated computational fluid dynamics could address the computational challenges in hydro-morpho-dynamic modeling, allowing for more precise predictions of stream flow and flood levels, particularly in ungauged basins \cite{kochkov2021machine}.

Additionally, AI can aid in calibrating non-contact video gauges, potentially more robust than traditional methods, as they are not directly affected by the water. AI can also guide forensic analysis to evaluate exposure and vulnerability, employing multi-modal approaches to refine geospatial models at a local scale. However, the effectiveness of these AI-driven models depends on detailed knowledge of the local terrain, including potential bottlenecks like bridges and channels, and accurate data about societal vulnerability to floods.

Yet, AI can also transform how warnings are issued to improve communication and response strategies. For instance, AI-generated maps and photorealistic visualizations based on digital elevation models can depict expected inundation areas and damages \cite{lutjens2023physicallyconsistent}. 
Moreover, AI can generate easily understandable, language-based warnings, both written and auditory, tailored for diverse populations, including the visually impaired. An LLM-based chatbot feature could enhance interactivity, providing real-time, personalized responses to emergency inquiries (see examples in Fig.~\ref{fig:casestudies}d).

\section{Conclusions and Outlook} 
\label{sec:conclusions}

This review highlighted the significant potential of AI for analyzing and modeling extreme events while also detailing the main difficulties and prospects associated with this emerging field. 
Integrating AI into extreme event analysis faces several challenges, including data management issues like handling dynamic datasets, biases, and high dimensionality that complicate feature extraction. AI models also struggle with unclear statistical definitions of what is `extreme'. 
Furthermore, integrating AI with physical models poses substantial challenges yet offers promising opportunities for enhancing model accuracy and reliability.
Trustworthiness concerns arise from the complexity and interpretability of ML models, the difficulty of generalizing across different contexts, and the quantification of uncertainty. 
Operational challenges include the complexity of AI outputs which hinder interpretation by non-experts, resistance to AI adoption due to concerns over reliability and fairness, and the need for frameworks that facilitate transparent and ethical integration of AI insights into decision-making processes.

The previous challenges compromise reproducibility and comparability of ML models in analyzing extreme events. 
These challenges are exacerbated by data scarcity, lack of transparency in model configurations, and the use of proprietary tools. Additionally, interdisciplinary differences hinder consistency and comparability. Effective solutions demand robust, transparent methodologies, inclusive data-sharing practices, and frameworks that support cross-disciplinary collaboration.

We overviewed and emphasized the importance of developing operational, explainable, and trustworthy AI systems. 
Addressing these challenges requires a coordinated effort across disciplines involving AI researchers, environmental and climate scientists, field experts, and policymakers. This collaborative approach is crucial for advancing AI applications in extreme event analysis and ensuring that these technologies are adapted to real-world needs and constraints. 
From an operational perspective, adapting AI solutions to real-time data integration, model deployment, and resource allocation highlights the need for systems that can function within the operational frameworks of disaster management and risk mitigation. Besides, methodological improvements are still needed in model evaluation and benchmarking to alleviate issues like overfitting and enhance AI systems' generalization capabilities. 

Looking forward, there are significant areas for further exploration and improvement. These include the development of benchmarks specific to extreme events, enhanced integration of domain knowledge to improve data fusion and model training, and the creation of robust, scalable AI systems capable of adapting to the dynamic nature of extreme events. Recent LLMs harvest vast amounts of domain knowledge embedded in the literature, promising significant advances in communicating risk.

As we advance, the ultimate goal is to harness AI's potential to substantially benefit society, particularly by enhancing our capacity to manage and respond to extreme events. Through dedicated research and collaborative innovation, AI can become a cornerstone in our strategy to understand and mitigate the impacts of these challenging and elusive phenomena.


\begin{thebibliography}{110}
\ifx \bisbn   \undefined \def \bisbn  #1{ISBN #1}\fi
\ifx \binits  \undefined \def \binits#1{#1}\fi
\ifx \bauthor  \undefined \def \bauthor#1{#1}\fi
\ifx \batitle  \undefined \def \batitle#1{#1}\fi
\ifx \bjtitle  \undefined \def \bjtitle#1{#1}\fi
\ifx \bvolume  \undefined \def \bvolume#1{\textbf{#1}}\fi
\ifx \byear  \undefined \def \byear#1{#1}\fi
\ifx \bissue  \undefined \def \bissue#1{#1}\fi
\ifx \bfpage  \undefined \def \bfpage#1{#1}\fi
\ifx \blpage  \undefined \def \blpage #1{#1}\fi
\ifx \burl  \undefined \def \burl#1{\textsf{#1}}\fi
\ifx \doiurl  \undefined \def \doiurl#1{\url{https://doi.org/#1}}\fi
\ifx \betal  \undefined \def \betal{\textit{et al.}}\fi
\ifx \binstitute  \undefined \def \binstitute#1{#1}\fi
\ifx \binstitutionaled  \undefined \def \binstitutionaled#1{#1}\fi
\ifx \bctitle  \undefined \def \bctitle#1{#1}\fi
\ifx \beditor  \undefined \def \beditor#1{#1}\fi
\ifx \bpublisher  \undefined \def \bpublisher#1{#1}\fi
\ifx \bbtitle  \undefined \def \bbtitle#1{#1}\fi
\ifx \bedition  \undefined \def \bedition#1{#1}\fi
\ifx \bseriesno  \undefined \def \bseriesno#1{#1}\fi
\ifx \blocation  \undefined \def \blocation#1{#1}\fi
\ifx \bsertitle  \undefined \def \bsertitle#1{#1}\fi
\ifx \bsnm \undefined \def \bsnm#1{#1}\fi
\ifx \bsuffix \undefined \def \bsuffix#1{#1}\fi
\ifx \bparticle \undefined \def \bparticle#1{#1}\fi
\ifx \barticle \undefined \def \barticle#1{#1}\fi
\bibcommenthead
\ifx \bconfdate \undefined \def \bconfdate #1{#1}\fi
\ifx \botherref \undefined \def \botherref #1{#1}\fi
\ifx \url \undefined \def \url#1{\textsf{#1}}\fi
\ifx \bchapter \undefined \def \bchapter#1{#1}\fi
\ifx \bbook \undefined \def \bbook#1{#1}\fi
\ifx \bcomment \undefined \def \bcomment#1{#1}\fi
\ifx \oauthor \undefined \def \oauthor#1{#1}\fi
\ifx \citeauthoryear \undefined \def \citeauthoryear#1{#1}\fi
\ifx \endbibitem  \undefined \def \endbibitem {}\fi
\ifx \bconflocation  \undefined \def \bconflocation#1{#1}\fi
\ifx \arxivurl  \undefined \def \arxivurl#1{\textsf{#1}}\fi
\csname PreBibitemsHook\endcsname

\bibitem{seneviratne2021weather}
\begin{botherref}
\oauthor{\bsnm{Seneviratne}, \binits{S.I.}},
\oauthor{\bsnm{Zhang}, \binits{X.}},
\oauthor{\bsnm{Adnan}, \binits{M.}},
\oauthor{\bsnm{Badi}, \binits{W.}},
\oauthor{\bsnm{Dereczynski}, \binits{C.}},
\oauthor{\bsnm{Di~Luca}, \binits{A.}},
\oauthor{\bsnm{Ghosh}, \binits{S.}},
\oauthor{\bsnm{Iskander}, \binits{I.}},
\oauthor{\bsnm{Kossin}, \binits{J.}},
\oauthor{\bsnm{Lewis}, \binits{S.}}, et al.:
Weather and climate extreme events in a changing climate (chapter 11)
(2021)
\end{botherref}
\endbibitem

\bibitem{aidetoolbox}
\begin{barticle}
\bauthor{\bsnm{Gonzalez-Calabuig}, \binits{M.}},
\bauthor{\bsnm{Cort{\'e}s-Andr{\'e}s}, \binits{J.}},
\bauthor{\bsnm{Williams}, \binits{T.}},
\bauthor{\bsnm{Zhang}, \binits{M.}},
\bauthor{\bsnm{Pellicer-Valero}, \binits{O.J.}},
\bauthor{\bsnm{Fern{\'a}ndez-Torres}, \binits{M.-{\'A}.}},
\bauthor{\bsnm{Camps-Valls}, \binits{G.}}:
\batitle{{The AIDE Toolbox: AI for Disentangling Extreme Events}}.
\bjtitle{IEEE Geoscience and Remote Sensing Magazine}
\bvolume{12}(\bissue{3}),
\bfpage{1}--\blpage{8}
(\byear{2024}).
\doiurl{10.1109/MGRS.2024.3382544}
\end{barticle}
\endbibitem

\bibitem{Lam2023}
\begin{barticle}
\bauthor{\bsnm{Lam}, \binits{R.}},
\bauthor{\bsnm{Sanchez-Gonzalez}, \binits{A.}},
\bauthor{\bsnm{Willson}, \binits{M.}},
\bauthor{\bsnm{Wirnsberger}, \binits{P.}},
\bauthor{\bsnm{Fortunato}, \binits{M.}},
\bauthor{\bsnm{Alet}, \binits{F.}},
\bauthor{\bsnm{Ravuri}, \binits{S.}},
\bauthor{\bsnm{Ewalds}, \binits{T.}},
\bauthor{\bsnm{Eaton-Rosen}, \binits{Z.}},
\bauthor{\bsnm{Hu}, \binits{W.}},
\bauthor{\bsnm{Merose}, \binits{A.}},
\bauthor{\bsnm{Hoyer}, \binits{S.}},
\bauthor{\bsnm{Holland}, \binits{G.}},
\bauthor{\bsnm{Vinyals}, \binits{O.}},
\bauthor{\bsnm{Stott}, \binits{J.}},
\bauthor{\bsnm{Pritzel}, \binits{A.}},
\bauthor{\bsnm{Mohamed}, \binits{S.}},
\bauthor{\bsnm{Battaglia}, \binits{P.}}:
\batitle{Learning skillful medium-range global weather forecasting}.
\bjtitle{Science}
\bvolume{382},
\bfpage{1416}--\blpage{1422}
(\byear{2023})
\end{barticle}
\endbibitem

\bibitem{ferchichi2022}
\begin{botherref}
\oauthor{\bsnm{Ferchichi}, \binits{A.}},
\oauthor{\bsnm{Abbes}, \binits{A.B.}},
\oauthor{\bsnm{Barra}, \binits{V.}},
\oauthor{\bsnm{Farah}, \binits{I.R.}}:
Forecasting vegetation indices from spatio-temporal remotely sensed data using
  deep learning-based approaches: {{A}} systematic literature review.
Ecological Informatics,
101552
(2022).
\doiurl{10.1016/j.ecoinf.2022.101552}
\end{botherref}
\endbibitem

\bibitem{ragone2021rare}
\begin{barticle}
\bauthor{\bsnm{Ragone}, \binits{F.}},
\bauthor{\bsnm{Bouchet}, \binits{F.}}:
\batitle{Rare event algorithm study of extreme warm summers and heatwaves over
  europe}.
\bjtitle{Geophysical Research Letters}
\bvolume{48}(\bissue{12}),
\bfpage{2020}--\blpage{091197}
(\byear{2021})
\end{barticle}
\endbibitem

\bibitem{Yokota2004}
\begin{barticle}
\bauthor{\bsnm{Yokota}, \binits{F.}},
\bauthor{\bsnm{Thompson}, \binits{K.M.}}:
\batitle{Value of information analysis in environmental health risk management
  decisions: past, present, and future}.
\bjtitle{Risk analysis}
\bvolume{24}(\bissue{3}),
\bfpage{635}--\blpage{650}
(\byear{2004})
\end{barticle}
\endbibitem

\bibitem{salcedosanz2024analysis}
\begin{barticle}
\bauthor{\bsnm{Salcedo-Sanz}, \binits{S.}},
\bauthor{\bsnm{P{\'{e}}rez-Aracil}, \binits{J.}},
\bauthor{\bsnm{Ascenso}, \binits{G.}},
\bauthor{\bsnm{{Del Ser}}, \binits{J.}},
\bauthor{\bsnm{Casillas-P{\'{e}}rez}, \binits{D.}},
\bauthor{\bsnm{Kadow}, \binits{C.}},
\bauthor{\bsnm{Fister}, \binits{D.}},
\bauthor{\bsnm{Barriopedro}, \binits{D.}},
\bauthor{\bsnm{Garc{\'{i}}a-Herrera}, \binits{R.}},
\bauthor{\bsnm{Giuliani}, \binits{M.}},
\bauthor{\bsnm{Castelletti}, \binits{A.}}:
\batitle{{Analysis, characterization, prediction, and attribution of extreme
  atmospheric events with machine learning and deep learning techniques: a
  review}}.
\bjtitle{Theoretical and Applied Climatology}
\bvolume{155}(\bissue{1}),
\bfpage{1}--\blpage{44}
(\byear{2024}).
\doiurl{10.1007/s00704-023-04571-5}
\end{barticle}
\endbibitem

\bibitem{gawlikowski2023survey}
\begin{barticle}
\bauthor{\bsnm{Gawlikowski}, \binits{J.}},
\bauthor{\bsnm{Tassi}, \binits{C.R.N.}},
\bauthor{\bsnm{Ali}, \binits{M.}},
\bauthor{\bsnm{Lee}, \binits{J.}},
\bauthor{\bsnm{Humt}, \binits{M.}},
\bauthor{\bsnm{Feng}, \binits{J.}},
\bauthor{\bsnm{Kruspe}, \binits{A.}},
\bauthor{\bsnm{Triebel}, \binits{R.}},
\bauthor{\bsnm{Jung}, \binits{P.}},
\bauthor{\bsnm{Roscher}, \binits{R.}},
\bauthor{\bsnm{Shahzad}, \binits{M.}},
\bauthor{\bsnm{Yang}, \binits{W.}},
\bauthor{\bsnm{Bamler}, \binits{R.}},
\bauthor{\bsnm{Zhu}, \binits{X.X.}}:
\batitle{{A survey of uncertainty in deep neural networks}}.
\bjtitle{Artificial Intelligence Review}
\bvolume{56}(\bissue{Suppl 1}),
\bfpage{1513}--\blpage{1589}
(\byear{2023})
{\href{https://arxiv.org/abs/2107.03342}{{arXiv:2107.03342}}}.
\doiurl{10.1007/s10462-023-10562-9}
\end{barticle}
\endbibitem

\bibitem{harrington2021quantifying}
\begin{barticle}
\bauthor{\bsnm{Harrington}, \binits{L.J.}},
\bauthor{\bsnm{Schleussner}, \binits{C.F.}},
\bauthor{\bsnm{Otto}, \binits{F.E.L.}}:
\batitle{{Quantifying uncertainty in aggregated climate change risk
  assessments}}.
\bjtitle{Nature Communications}
\bvolume{12}(\bissue{1}),
\bfpage{7140}
(\byear{2021}).
\doiurl{10.1038/s41467-021-27491-2}
\end{barticle}
\endbibitem

\bibitem{stott2016attribution}
\begin{barticle}
\bauthor{\bsnm{Stott}, \binits{P.A.}},
\bauthor{\bsnm{Christidis}, \binits{N.}},
\bauthor{\bsnm{Otto}, \binits{F.E.L.}},
\bauthor{\bsnm{Sun}, \binits{Y.}},
\bauthor{\bsnm{Vanderlinden}, \binits{J.P.}},
\bauthor{\bparticle{van} \bsnm{Oldenborgh}, \binits{G.J.}},
\bauthor{\bsnm{Vautard}, \binits{R.}},
\bauthor{\bparticle{von} \bsnm{Storch}, \binits{H.}},
\bauthor{\bsnm{Walton}, \binits{P.}},
\bauthor{\bsnm{Yiou}, \binits{P.}},
\bauthor{\bsnm{Zwiers}, \binits{F.W.}}:
\batitle{{Attribution of extreme weather and climate-related events}}.
\bjtitle{Wiley Interdisciplinary Reviews: Climate Change}
\bvolume{7}(\bissue{1}),
\bfpage{23}--\blpage{41}
(\byear{2016}).
\doiurl{10.1002/wcc.380}
\end{barticle}
\endbibitem

\bibitem{madakumbura2021anthropogenic}
\begin{barticle}
\bauthor{\bsnm{Madakumbura}, \binits{G.D.}},
\bauthor{\bsnm{Thackeray}, \binits{C.W.}},
\bauthor{\bsnm{Norris}, \binits{J.}},
\bauthor{\bsnm{Goldenson}, \binits{N.}},
\bauthor{\bsnm{Hall}, \binits{A.}}:
\batitle{Anthropogenic influence on extreme precipitation over global land
  areas seen in multiple observational datasets}.
\bjtitle{Nature Communications}
\bvolume{12}(\bissue{1}),
\bfpage{3944}
(\byear{2021})
\end{barticle}
\endbibitem

\bibitem{chattopadhyay_analog_2020}
\begin{botherref}
\oauthor{\bsnm{Chattopadhyay}, \binits{A.}},
\oauthor{\bsnm{Nabizadeh}, \binits{E.}},
\oauthor{\bsnm{Hassanzadeh}, \binits{P.}}:
Analog forecasting of extreme-causing weather patterns using deep learning.
Journal of Advances in Modeling {E}arth Systems
\textbf{12}(2),
2019--001958.
\doiurl{10.1029/2019MS001958}
\end{botherref}
\endbibitem

\bibitem{FindingtherightXAIMethodBommer2024}
\begin{barticle}
\bauthor{\bsnm{Bommer}, \binits{P.L.}},
\bauthor{\bsnm{Kretschmer}, \binits{M.}},
\bauthor{\bsnm{Hedstrom}, \binits{A.}},
\bauthor{\bsnm{Bareeva}, \binits{D.}},
\bauthor{\bsnm{Hohne}, \binits{M.M.-C.}}:
\batitle{Finding the right {XAI} method — {A} guide for the evaluation and
  ranking of explainable ai methods in climate science}.
\bjtitle{Artificial Intelligence for the {E}arth Systems}
(\byear{2024}).
\doiurl{10.1175/AIES-D-23-0074.1}
\end{barticle}
\endbibitem

\bibitem{Hannart2016CausalCT}
\begin{barticle}
\bauthor{\bsnm{Hannart}, \binits{A.}},
\bauthor{\bsnm{Pearl}, \binits{J.}},
\bauthor{\bsnm{Otto}, \binits{F.E.L.}},
\bauthor{\bsnm{Naveau}, \binits{P.}},
\bauthor{\bsnm{Ghil}, \binits{M.}}:
\batitle{Causal counterfactual theory for the attribution of weather and
  climate-related events}.
\bjtitle{Bulletin of the American Meteorological Society}
\bvolume{97},
\bfpage{99}--\blpage{110}
(\byear{2016})
\end{barticle}
\endbibitem

\bibitem{vaghefi2023chatclimate}
\begin{barticle}
\bauthor{\bsnm{Vaghefi}, \binits{S.A.}},
\bauthor{\bsnm{Stammbach}, \binits{D.}},
\bauthor{\bsnm{Muccione}, \binits{V.}},
\bauthor{\bsnm{Bingler}, \binits{J.}},
\bauthor{\bsnm{Ni}, \binits{J.}},
\bauthor{\bsnm{Kraus}, \binits{M.}},
\bauthor{\bsnm{Allen}, \binits{S.}},
\bauthor{\bsnm{Colesanti-Senni}, \binits{C.}},
\bauthor{\bsnm{Wekhof}, \binits{T.}},
\bauthor{\bsnm{Schimanski}, \binits{T.}}, \betal:
\batitle{{ChatClimate: Grounding conversational AI in climate science}}.
\bjtitle{Communications {E}arth \& Environment}
\bvolume{4}(\bissue{1}),
\bfpage{480}
(\byear{2023})
\end{barticle}
\endbibitem

\bibitem{ruff2021unifying}
\begin{barticle}
\bauthor{\bsnm{Ruff}, \binits{L.}},
\bauthor{\bsnm{Kauffmann}, \binits{J.R.}},
\bauthor{\bsnm{Vandermeulen}, \binits{R.A.}},
\bauthor{\bsnm{Montavon}, \binits{G.}},
\bauthor{\bsnm{Samek}, \binits{W.}},
\bauthor{\bsnm{Kloft}, \binits{M.}},
\bauthor{\bsnm{Dietterich}, \binits{T.G.}},
\bauthor{\bsnm{M{u}ller}, \binits{K.-R.}}:
\batitle{A unifying review of deep and shallow anomaly detection}.
\bjtitle{Proceedings of the IEEE}
\bvolume{109}(\bissue{5}),
\bfpage{756}--\blpage{795}
(\byear{2021})
\end{barticle}
\endbibitem

\bibitem{han2019review}
\begin{barticle}
\bauthor{\bsnm{Han}, \binits{Z.}},
\bauthor{\bsnm{Zhao}, \binits{J.}},
\bauthor{\bsnm{Leung}, \binits{H.}},
\bauthor{\bsnm{Ma}, \binits{K.F.}},
\bauthor{\bsnm{Wang}, \binits{W.}}:
\batitle{A review of deep learning models for time series prediction}.
\bjtitle{IEEE Sensors Journal}
\bvolume{21}(\bissue{6}),
\bfpage{7833}--\blpage{7848}
(\byear{2019})
\end{barticle}
\endbibitem

\bibitem{zennaro2021exploring}
\begin{barticle}
\bauthor{\bsnm{Zennaro}, \binits{F.}},
\bauthor{\bsnm{Furlan}, \binits{E.}},
\bauthor{\bsnm{Simeoni}, \binits{C.}},
\bauthor{\bsnm{Torresan}, \binits{S.}},
\bauthor{\bsnm{Aslan}, \binits{S.}},
\bauthor{\bsnm{Critto}, \binits{A.}},
\bauthor{\bsnm{Marcomini}, \binits{A.}}:
\batitle{{Exploring machine learning potential for climate change risk
  assessment}}.
\bjtitle{Earth-Science Reviews}
\bvolume{220},
\bfpage{103752}
(\byear{2021}).
\doiurl{10.1016/j.earscirev.2021.103752}
\end{barticle}
\endbibitem

\bibitem{Jones_2023}
\begin{barticle}
\bauthor{\bsnm{Jones}, \binits{R.L.}},
\bauthor{\bsnm{Kharb}, \binits{A.}},
\bauthor{\bsnm{Tubeuf}, \binits{S.}}:
\batitle{The untold story of missing data in disaster research: a systematic
  review of the empirical literature utilising the emergency events database
  ({EM-DAT})}.
\bjtitle{Environmental Research Letters}
\bvolume{18}(\bissue{10}),
\bfpage{103006}
(\byear{2023}).
\doiurl{10.1088/1748-9326/acfd42}
\end{barticle}
\endbibitem

\bibitem{flach2017multivariate}
\begin{barticle}
\bauthor{\bsnm{Flach}, \binits{M.}},
\bauthor{\bsnm{Gans}, \binits{F.}},
\bauthor{\bsnm{Brenning}, \binits{A.}},
\bauthor{\bsnm{Denzler}, \binits{J.}},
\bauthor{\bsnm{Reichstein}, \binits{M.}},
\bauthor{\bsnm{Rodner}, \binits{E.}},
\bauthor{\bsnm{Bathiany}, \binits{S.}},
\bauthor{\bsnm{Bodesheim}, \binits{P.}},
\bauthor{\bsnm{Guanche}, \binits{Y.}},
\bauthor{\bsnm{Sippel}, \binits{S.}}, \betal:
\batitle{Multivariate anomaly detection for {E}arth observations: a comparison
  of algorithms and feature extraction techniques}.
\bjtitle{Earth System Dynamics}
\bvolume{8}(\bissue{3}),
\bfpage{677}--\blpage{696}
(\byear{2017})
\end{barticle}
\endbibitem

\bibitem{karaismailoglu2021climatenet}
\begin{barticle}
\bauthor{\bsnm{Prabhat}},
\bauthor{\bsnm{Kashinath}, \binits{K.}},
\bauthor{\bsnm{Mudigonda}, \binits{M.}},
\bauthor{\bsnm{Kim}, \binits{S.}},
\bauthor{\bsnm{Kapp-Schwoerer}, \binits{L.}},
\bauthor{\bsnm{Graubner}, \binits{A.}},
\bauthor{\bsnm{Karaismailoglu}, \binits{E.}},
\bauthor{\bparticle{von} \bsnm{Kleist}, \binits{L.}},
\bauthor{\bsnm{Kurth}, \binits{T.}},
\bauthor{\bsnm{Greiner}, \binits{A.}}, \betal:
\batitle{{ClimateNet: An expert-labelled open dataset and Deep Learning
  architecture for enabling high-precision analyses of extreme weather}}.
\bjtitle{Geoscientific Model Development Discussions}
\bvolume{2020},
\bfpage{1}--\blpage{28}
(\byear{2020})
\end{barticle}
\endbibitem

\bibitem{racah2017extremeweather}
\begin{botherref}
\oauthor{\bsnm{Racah}, \binits{E.}},
\oauthor{\bsnm{Beckham}, \binits{C.}},
\oauthor{\bsnm{Maharaj}, \binits{T.}},
\oauthor{\bsnm{Ebrahimi~Kahou}, \binits{S.}},
\oauthor{\bsnm{Prabhat}, \binits{M.}},
\oauthor{\bsnm{Pal}, \binits{C.}}:
Extremeweather: A large-scale climate dataset for semi-supervised detection,
  localization, and understanding of extreme weather events.
Advances in Neural Information Processing Systems (NeurIPS)
\textbf{30}
(2017)
\end{botherref}
\endbibitem

\bibitem{GuancheGarca2018}
\begin{barticle}
\bauthor{\bsnm{Guanche~Garc\'ia}, \binits{Y.}},
\bauthor{\bsnm{Shadaydeh}, \binits{M.}},
\bauthor{\bsnm{Mahecha}, \binits{M.}},
\bauthor{\bsnm{Denzler}, \binits{J.}}:
\batitle{Extreme anomaly event detection in biosphere using linear regression
  and a spatiotemporal mrf model}.
\bjtitle{Natural Hazards}
\bvolume{98}(\bissue{3}),
\bfpage{849}--\blpage{867}
(\byear{2018}).
\doiurl{10.1007/s11069-018-3415-8}
\end{barticle}
\endbibitem

\bibitem{Klehmet2024}
\begin{barticle}
\bauthor{\bsnm{Klehmet}, \binits{K.}},
\bauthor{\bsnm{Berg}, \binits{P.}},
\bauthor{\bsnm{Bozhinova}, \binits{D.}},
\bauthor{\bsnm{Crochemore}, \binits{L.}},
\bauthor{\bsnm{Du}, \binits{Y.}},
\bauthor{\bsnm{Pechlivanidis}, \binits{I.}},
\bauthor{\bsnm{Photiadou}, \binits{C.}},
\bauthor{\bsnm{Yang}, \binits{W.}}:
\batitle{Robustness of hydrometeorological extremes in surrogated seasonal
  forecasts}.
\bjtitle{International Journal of Climatology}
\bvolume{44}(\bissue{5}),
\bfpage{1725}--\blpage{1738}
(\byear{2024}).
\doiurl{10.1002/joc.8407}
\end{barticle}
\endbibitem

\bibitem{Johnson20sakame}
\begin{barticle}
\bauthor{\bsnm{Johnson}, \binits{J.E.}},
\bauthor{\bsnm{Laparra}, \binits{V.}},
\bauthor{\bsnm{P\'{e}rez-Suay}, \binits{A.}},
\bauthor{\bsnm{Mahecha}, \binits{M.}},
\bauthor{\bsnm{Camps-Valls}, \binits{G.}}:
\batitle{Kernel methods and their derivatives: Concept and perspectives for the
  {E}arth system sciences}.
\bjtitle{PLOS One}
\bvolume{15}(\bissue{10}),
\bfpage{0235885}
(\byear{2020})
\end{barticle}
\endbibitem

\bibitem{Johnson20rbig4eo}
\begin{barticle}
\bauthor{\bsnm{Johnson}, \binits{J.E.}},
\bauthor{\bsnm{Laparra}, \binits{V.}},
\bauthor{\bsnm{Piles}, \binits{M.}},
\bauthor{\bsnm{Camps-Valls}, \binits{G.}}:
\batitle{Gaussianizing the {E}arth: Multidimensional information measures for
  {E}arth data analysis}.
\bjtitle{IEEE Geoscience and Remote Sensing Magazine}
\bvolume{9}(\bissue{4}),
\bfpage{191}--\blpage{208}
(\byear{2021}).
\doiurl{10.1109/MGRS.2021.3066260}
\end{barticle}
\endbibitem

\bibitem{vijverberg2022role}
\begin{barticle}
\bauthor{\bsnm{Vijverberg}, \binits{S.}},
\bauthor{\bsnm{Coumou}, \binits{D.}}:
\batitle{{The role of the Pacific Decadal Oscillation and ocean-atmosphere
  interactions in driving US temperature variability}}.
\bjtitle{npj Climate and Atmospheric Science}
\bvolume{5}(\bissue{1}),
\bfpage{18}
(\byear{2022}).
\doiurl{10.1038/s41612-022-00237-7}
\end{barticle}
\endbibitem

\bibitem{kladny2024enhanced}
\begin{botherref}
\oauthor{\bsnm{Kladny}, \binits{K.-R.}},
\oauthor{\bsnm{Milanta}, \binits{M.}},
\oauthor{\bsnm{Mraz}, \binits{O.}},
\oauthor{\bsnm{Hufkens}, \binits{K.}},
\oauthor{\bsnm{Stocker}, \binits{B.D.}}:
Enhanced prediction of vegetation responses to extreme drought using deep
  learning and {E}arth observation data.
Ecological Informatics,
102474
(2024)
\end{botherref}
\endbibitem

\bibitem{bentivoglio2022deep}
\begin{barticle}
\bauthor{\bsnm{Bentivoglio}, \binits{R.}},
\bauthor{\bsnm{Isufi}, \binits{E.}},
\bauthor{\bsnm{Jonkman}, \binits{S.N.}},
\bauthor{\bsnm{Taormina}, \binits{R.}}:
\batitle{Deep learning methods for flood mapping: a review of existing
  applications and future research directions}.
\bjtitle{Hydrology and {E}arth system sciences}
\bvolume{26}(\bissue{16}),
\bfpage{4345}--\blpage{4378}
(\byear{2022})
\end{barticle}
\endbibitem

\bibitem{belayneh2014long}
\begin{barticle}
\bauthor{\bsnm{Belayneh}, \binits{A.}},
\bauthor{\bsnm{Adamowski}, \binits{J.}},
\bauthor{\bsnm{Khalil}, \binits{B.}},
\bauthor{\bsnm{Ozga-Zielinski}, \binits{B.}}:
\batitle{{Long-term SPI drought forecasting in the Awash River Basin in
  Ethiopia using wavelet neural network and wavelet support vector regression
  models}}.
\bjtitle{Journal of Hydrology}
\bvolume{508},
\bfpage{418}--\blpage{429}
(\byear{2014})
\end{barticle}
\endbibitem

\bibitem{kondylatos_wildfire_2022}
\begin{barticle}
\bauthor{\bsnm{Kondylatos}, \binits{S.}},
\bauthor{\bsnm{Prapas}, \binits{I.}},
\bauthor{\bsnm{Ronco}, \binits{M.}},
\bauthor{\bsnm{Papoutsis}, \binits{I.}},
\bauthor{\bsnm{Camps-Valls}, \binits{G.}},
\bauthor{\bsnm{Piles}, \binits{M.}},
\bauthor{\bsnm{Fern\'andez-Torres}, \binits{M.-A.}},
\bauthor{\bsnm{Carvalhais}, \binits{N.}}:
\batitle{Wildfire {Danger} {Prediction} and {Understanding} {With} {Deep}
  {Learning}}.
\bjtitle{Geophysical Research Letters}
\bvolume{49}(\bissue{17}),
\bfpage{2022}--\blpage{099368}
(\byear{2022}).
\doiurl{10.1029/2022GL099368}
\end{barticle}
\endbibitem

\bibitem{nearing2024global}
\begin{barticle}
\bauthor{\bsnm{Nearing}, \binits{G.}},
\bauthor{\bsnm{Cohen}, \binits{D.}},
\bauthor{\bsnm{Dube}, \binits{V.}},
\bauthor{\bsnm{Gauch}, \binits{M.}},
\bauthor{\bsnm{Gilon}, \binits{O.}},
\bauthor{\bsnm{Harrigan}, \binits{S.}},
\bauthor{\bsnm{Hassidim}, \binits{A.}},
\bauthor{\bsnm{Klotz}, \binits{D.}},
\bauthor{\bsnm{Kratzert}, \binits{F.}},
\bauthor{\bsnm{Metzger}, \binits{A.}}, \betal:
\batitle{Global prediction of extreme floods in ungauged watersheds}.
\bjtitle{Nature}
\bvolume{627}(\bissue{8004}),
\bfpage{559}--\blpage{563}
(\byear{2024})
\end{barticle}
\endbibitem

\bibitem{zhang2021deep}
\begin{barticle}
\bauthor{\bsnm{Zhang}, \binits{G.}},
\bauthor{\bsnm{Wang}, \binits{M.}},
\bauthor{\bsnm{Liu}, \binits{K.}}:
\batitle{Deep neural networks for global wildfire susceptibility modelling}.
\bjtitle{Ecological Indicators}
\bvolume{127},
\bfpage{107735}
(\byear{2021})
\end{barticle}
\endbibitem

\bibitem{miloshevich2023probabilistic}
\begin{barticle}
\bauthor{\bsnm{Miloshevich}, \binits{G.}},
\bauthor{\bsnm{Cozian}, \binits{B.}},
\bauthor{\bsnm{Abry}, \binits{P.}},
\bauthor{\bsnm{Borgnat}, \binits{P.}},
\bauthor{\bsnm{Bouchet}, \binits{F.}}:
\batitle{{Probabilistic forecasts of extreme heatwaves using convolutional
  neural networks in a regime of lack of data}}.
\bjtitle{Physical Review Fluids}
\bvolume{8}(\bissue{4}),
\bfpage{40501}
(\byear{2023})
{\href{https://arxiv.org/abs/2208.00971}{{arXiv:2208.00971}}}.
\doiurl{10.1103/PhysRevFluids.8.040501}
\end{barticle}
\endbibitem

\bibitem{vo2023lstm}
\begin{barticle}
\bauthor{\bsnm{Vo}, \binits{T.Q.}},
\bauthor{\bsnm{Kim}, \binits{S.-H.}},
\bauthor{\bsnm{Nguyen}, \binits{D.H.}},
\bauthor{\bsnm{Bae}, \binits{D.-H.}}:
\batitle{{LSTM-CM: a hybrid approach for natural drought prediction based on
  deep learning and climate models}}.
\bjtitle{Stochastic Environmental Research and Risk Assessment}
\bvolume{37}(\bissue{6}),
\bfpage{2035}--\blpage{2051}
(\byear{2023})
\end{barticle}
\endbibitem

\bibitem{shi2020enabling}
\begin{barticle}
\bauthor{\bsnm{Shi}, \binits{X.}}:
\batitle{Enabling smart dynamical downscaling of extreme precipitation events
  with machine learning}.
\bjtitle{Geophysical Research Letters}
\bvolume{47}(\bissue{19}),
\bfpage{2020}--\blpage{090309}
(\byear{2020})
\end{barticle}
\endbibitem

\bibitem{callaghan2021machine}
\begin{barticle}
\bauthor{\bsnm{Callaghan}, \binits{M.}},
\bauthor{\bsnm{Schleussner}, \binits{C.-F.}},
\bauthor{\bsnm{Nath}, \binits{S.}},
\bauthor{\bparticle{et} \bsnm{al.}}:
\batitle{Machine-learning-based evidence and attribution mapping of 100,000
  climate impact studies}.
\bjtitle{Nature Climate Change}
\bvolume{11},
\bfpage{966}--\blpage{972}
(\byear{2021}).
\doiurl{10.1038/s41558-021-01168-6}
\end{barticle}
\endbibitem

\bibitem{sutanto2019moving}
\begin{barticle}
\bauthor{\bsnm{Sutanto}, \binits{S.J.}},
\bauthor{\bparticle{van~der} \bsnm{Weert}, \binits{M.}},
\bauthor{\bsnm{Wanders}, \binits{N.}}, \betal:
\batitle{Moving from drought hazard to impact forecasts}.
\bjtitle{Nature communications}
\bvolume{10},
\bfpage{4945}
(\byear{2019}).
\doiurl{10.1038/s41467-019-12840-z}
\end{barticle}
\endbibitem

\bibitem{salakpi2022forecasting}
\begin{barticle}
\bauthor{\bsnm{Salakpi}, \binits{E.E.}},
\bauthor{\bsnm{Hurley}, \binits{P.D.}},
\bauthor{\bsnm{Muthoka}, \binits{J.M.}},
\bauthor{\bsnm{Barrett}, \binits{A.B.}},
\bauthor{\bsnm{Bowell}, \binits{A.}},
\bauthor{\bsnm{Oliver}, \binits{S.}},
\bauthor{\bsnm{Rowhani}, \binits{P.}}:
\batitle{{Forecasting vegetation condition with a Bayesian auto-regressive
  distributed lags (BARDL) model}}.
\bjtitle{Natural Hazards and {E}arth System Sciences}
\bvolume{22}(\bissue{8}),
\bfpage{2703}--\blpage{2723}
(\byear{2022})
\end{barticle}
\endbibitem

\bibitem{martinuzzi2023learning}
\begin{barticle}
\bauthor{\bsnm{Martinuzzi}, \binits{F.}},
\bauthor{\bsnm{Mahecha}, \binits{M.D.}},
\bauthor{\bsnm{Camps-Valls}, \binits{G.}},
\bauthor{\bsnm{Montero}, \binits{D.}},
\bauthor{\bsnm{Williams}, \binits{T.}},
\bauthor{\bsnm{Mora}, \binits{K.}}:
\batitle{Learning extreme vegetation response to climate forcing: A comparison
  of recurrent neural network architectures}.
\bjtitle{Nonlinear Processes in Geophysics}
\bvolume{2024},
\bfpage{1}--\blpage{32}
(\byear{2024})
\end{barticle}
\endbibitem

\bibitem{ahmad2023machine}
\begin{barticle}
\bauthor{\bsnm{Ahmad}, \binits{R.}},
\bauthor{\bsnm{Yang}, \binits{B.}},
\bauthor{\bsnm{Ettlin}, \binits{G.}},
\bauthor{\bsnm{Berger}, \binits{A.}},
\bauthor{\bsnm{Rodr{\'\i}guez-Bocca}, \binits{P.}}:
\batitle{{A machine-learning based ConvLSTM architecture for NDVI
  forecasting}}.
\bjtitle{International Transactions in Operational Research}
\bvolume{30}(\bissue{4}),
\bfpage{2025}--\blpage{2048}
(\byear{2023})
\end{barticle}
\endbibitem

\bibitem{benson2023forecasting}
\begin{bchapter}
\bauthor{\bsnm{Benson}, \binits{V.}},
\bauthor{\bsnm{Requena-Mesa}, \binits{C.}},
\bauthor{\bsnm{Robin}, \binits{C.}},
\bauthor{\bsnm{Alonso}, \binits{L.}},
\bauthor{\bsnm{Cort{\'e}s}, \binits{J.}},
\bauthor{\bsnm{Gao}, \binits{Z.}},
\bauthor{\bsnm{Linscheid}, \binits{N.}},
\bauthor{\bsnm{Weynants}, \binits{M.}},
\bauthor{\bsnm{Reichstein}, \binits{M.}}:
\bctitle{Forecasting localized weather impacts on vegetation as seen from space
  with meteo-guided video prediction}.
In: \bbtitle{CVPR 2024}
(\byear{2024})
\end{bchapter}
\endbibitem

\bibitem{zhang2021impact}
\begin{barticle}
\bauthor{\bsnm{Zhang}, \binits{X.}},
\bauthor{\bsnm{Chen}, \binits{G.}},
\bauthor{\bsnm{Cai}, \binits{L.}},
\bauthor{\bsnm{Jiao}, \binits{H.}},
\bauthor{\bsnm{Hua}, \binits{J.}},
\bauthor{\bsnm{Luo}, \binits{X.}},
\bauthor{\bsnm{Wei}, \binits{X.}}:
\batitle{{Impact assessments of typhoon Lekima on forest damages in Subtropical
  China using machine learning methods and Landsat 8 OLI imagery}}.
\bjtitle{Sustainability}
\bvolume{13}(\bissue{9}),
\bfpage{4893}
(\byear{2021})
\end{barticle}
\endbibitem

\bibitem{ronco2023exploring}
\begin{barticle}
\bauthor{\bsnm{Ronco}, \binits{M.}},
\bauthor{\bsnm{T{\'a}rraga}, \binits{J.M.}},
\bauthor{\bsnm{Mu{\~n}oz}, \binits{J.}},
\bauthor{\bsnm{Piles}, \binits{M.}},
\bauthor{\bsnm{Marco}, \binits{E.S.}},
\bauthor{\bsnm{Wang}, \binits{Q.}},
\bauthor{\bsnm{Espinosa}, \binits{M.T.M.}},
\bauthor{\bsnm{Ponserre}, \binits{S.}},
\bauthor{\bsnm{Camps-Valls}, \binits{G.}}:
\batitle{Exploring interactions between socioeconomic context and natural
  hazards on human population displacement}.
\bjtitle{Nature Communications}
\bvolume{14}(\bissue{1}),
\bfpage{8004}
(\byear{2023})
\end{barticle}
\endbibitem

\bibitem{Sodoge2023}
\begin{barticle}
\bauthor{\bsnm{Sodoge}, \binits{J.}},
\bauthor{\bsnm{Kuhlicke}, \binits{C.}},
\bauthor{\bparticle{de} \bsnm{Brito}, \binits{M.M.}}:
\batitle{Automatized spatio-temporal detection of drought impacts from
  newspaper articles using natural language processing and machine learning}.
\bjtitle{Weather and Climate Extremes}
\bvolume{41},
\bfpage{100574}
(\byear{2023}).
\doiurl{10.1016/j.wace.2023.100574}
\end{barticle}
\endbibitem

\bibitem{bostromTrustTrustworthy}
\begin{botherref}
\oauthor{\bsnm{Bostrom}, \binits{A.}},
\oauthor{\bsnm{Demuth}, \binits{J.L.}},
\oauthor{\bsnm{Wirz}, \binits{C.D.}},
\oauthor{\bsnm{Cains}, \binits{M.G.}},
\oauthor{\bsnm{Schumacher}, \binits{A.}},
\oauthor{\bsnm{Madlambayan}, \binits{D.}},
\oauthor{\bsnm{Bansal}, \binits{A.S.}},
\oauthor{\bsnm{Bearth}, \binits{A.}},
\oauthor{\bsnm{Chase}, \binits{R.}},
\oauthor{\bsnm{Crosman}, \binits{K.M.}},
\oauthor{\bsnm{{Ebert-Uphoff}}, \binits{I.}},
\oauthor{\bsnm{Gagne~II}, \binits{D.J.}},
\oauthor{\bsnm{Guikema}, \binits{S.}},
\oauthor{\bsnm{Hoffman}, \binits{R.}},
\oauthor{\bsnm{Johnson}, \binits{B.B.}},
\oauthor{\bsnm{{Kumler-Bonfanti}}, \binits{C.}},
\oauthor{\bsnm{Lee}, \binits{J.D.}},
\oauthor{\bsnm{Lowe}, \binits{A.}},
\oauthor{\bsnm{McGovern}, \binits{A.}},
\oauthor{\bsnm{Przybylo}, \binits{V.}},
\oauthor{\bsnm{Radford}, \binits{J.T.}},
\oauthor{\bsnm{Roth}, \binits{E.}},
\oauthor{\bsnm{Sutter}, \binits{C.}},
\oauthor{\bsnm{Tissot}, \binits{P.}},
\oauthor{\bsnm{Roebber}, \binits{P.}},
\oauthor{\bsnm{Stewart}, \binits{J.Q.}},
\oauthor{\bsnm{White}, \binits{M.}},
\oauthor{\bsnm{Williams}, \binits{J.K.}}:
Trust and trustworthy artificial intelligence: {{A}} research agenda for {{AI}}
  in the environmental sciences.
Risk Analysis.
\doiurl{10.1111/risa.14245}
\end{botherref}
\endbibitem

\bibitem{naveau2020statistical}
\begin{barticle}
\bauthor{\bsnm{Naveau}, \binits{P.}},
\bauthor{\bsnm{Hannart}, \binits{A.}},
\bauthor{\bsnm{Ribes}, \binits{A.}}:
\batitle{{Statistical methods for extreme event attribution in climate
  science}}.
\bjtitle{Annual Review of Statistics and Its Application}
\bvolume{7},
\bfpage{89}--\blpage{110}
(\byear{2020}).
\doiurl{10.1146/annurev-statistics-031219-041314}
\end{barticle}
\endbibitem

\bibitem{otto2023attribution}
\begin{barticle}
\bauthor{\bsnm{Otto}, \binits{F.E.}}:
\batitle{Attribution of extreme events to climate change}.
\bjtitle{Annual Review of Environment and Resources}
\bvolume{48},
\bfpage{813}--\blpage{828}
(\byear{2023})
\end{barticle}
\endbibitem

\bibitem{pasini2017attribution}
\begin{barticle}
\bauthor{\bsnm{Pasini}, \binits{A.}},
\bauthor{\bsnm{Racca}, \binits{P.}},
\bauthor{\bsnm{Amendola}, \binits{S.}},
\bauthor{\bsnm{Cartocci}, \binits{G.}},
\bauthor{\bsnm{Cassardo}, \binits{C.}}:
\batitle{Attribution of recent temperature behaviour reassessed by a
  neural-network method}.
\bjtitle{Scientific reports}
\bvolume{7}(\bissue{1}),
\bfpage{17681}
(\byear{2017})
\end{barticle}
\endbibitem

\bibitem{barnes2019viewing}
\begin{barticle}
\bauthor{\bsnm{Barnes}, \binits{E.A.}},
\bauthor{\bsnm{Hurrell}, \binits{J.W.}},
\bauthor{\bsnm{Ebert-Uphoff}, \binits{I.}},
\bauthor{\bsnm{Anderson}, \binits{C.}},
\bauthor{\bsnm{Anderson}, \binits{D.}}:
\batitle{{Viewing forced climate patterns through an AI lens}}.
\bjtitle{Geophysical Research Letters}
\bvolume{46}(\bissue{22}),
\bfpage{13389}--\blpage{13398}
(\byear{2019})
\end{barticle}
\endbibitem

\bibitem{ghaffarianExplainableArtificial2023}
\begin{barticle}
\bauthor{\bsnm{Ghaffarian}, \binits{S.}},
\bauthor{\bsnm{Taghikhah}, \binits{F.R.}},
\bauthor{\bsnm{Maier}, \binits{H.R.}}:
\batitle{Explainable artificial intelligence in disaster risk management:
  {{Achievements}} and prospective futures}.
\bjtitle{International Journal of Disaster Risk Reduction}
\bvolume{98},
\bfpage{104123}
(\byear{2023}).
\doiurl{10.1016/j.ijdrr.2023.104123}
\end{barticle}
\endbibitem

\bibitem{Tuia24perspective}
\begin{botherref}
\oauthor{\bsnm{Tuia}, \binits{D.}},
\oauthor{\bsnm{Schindler}, \binits{K.}},
\oauthor{\bsnm{Demir}, \binits{B.}},
\oauthor{\bsnm{Camps-Valls}, \binits{G.}},
\oauthor{\bsnm{Zhu}, \binits{X.X.}},
\oauthor{\bsnm{Kochupillai}, \binits{M.}},
\oauthor{\bsnm{D\v{z}eroski}, \binits{S.}},
\oauthor{\bparticle{van} \bsnm{Rijn}, \binits{J.N.}},
\oauthor{\bsnm{Hoos}, \binits{H.H.}},
\oauthor{\bsnm{Frate}, \binits{F.D.}},
\oauthor{\bsnm{Datcu}, \binits{M.}},
\oauthor{\bsnm{Quian\'{e}-Ruiz}, \binits{J.-A.}},
\oauthor{\bsnm{Markl}, \binits{V.}},
\oauthor{\bsnm{Saux}, \binits{B.L.}},
\oauthor{\bsnm{Schneider}, \binits{R.}}:
Artificial intelligence to advance {E}arth observation: a perspective.
IEEE Geoscience and Remote Sensing Magazine
(2024)
\end{botherref}
\endbibitem

\bibitem{schlund2020constraining}
\begin{barticle}
\bauthor{\bsnm{Schlund}, \binits{M.}},
\bauthor{\bsnm{Eyring}, \binits{V.}},
\bauthor{\bsnm{Camps-Valls}, \binits{G.}},
\bauthor{\bsnm{Friedlingstein}, \binits{P.}},
\bauthor{\bsnm{Gentine}, \binits{P.}},
\bauthor{\bsnm{Reichstein}, \binits{M.}}:
\batitle{Constraining uncertainty in projected gross primary production with
  machine learning}.
\bjtitle{Journal of Geophysical Research: Biogeosciences}
\bvolume{125}(\bissue{11}),
\bfpage{2019}--\blpage{005619}
(\byear{2020})
\end{barticle}
\endbibitem

\bibitem{srinivasan2020machine}
\begin{barticle}
\bauthor{\bsnm{Srinivasan}, \binits{R.}},
\bauthor{\bsnm{Wang}, \binits{L.}},
\bauthor{\bsnm{Bulleid}, \binits{J.}}:
\batitle{Machine learning-based climate time series anomaly detection using
  convolutional neural networks}.
\bjtitle{Weather and Climate}
\bvolume{40}(\bissue{1}),
\bfpage{16}--\blpage{31}
(\byear{2020})
\end{barticle}
\endbibitem

\bibitem{dikshitSolvingTransparency2022a}
\begin{barticle}
\bauthor{\bsnm{Dikshit}, \binits{A.}},
\bauthor{\bsnm{Pradhan}, \binits{B.}},
\bauthor{\bsnm{Assiri}, \binits{M.E.}},
\bauthor{\bsnm{Almazroui}, \binits{M.}},
\bauthor{\bsnm{Park}, \binits{H.-J.}}:
\batitle{Solving transparency in drought forecasting using attention models}.
\bjtitle{Science of The Total Environment}
\bvolume{837},
\bfpage{155856}
(\byear{2022}).
\doiurl{10.1016/j.scitotenv.2022.155856}
\end{barticle}
\endbibitem

\bibitem{barnesThisLooks2022}
\begin{barticle}
\bauthor{\bsnm{Barnes}, \binits{E.A.}},
\bauthor{\bsnm{Barnes}, \binits{R.J.}},
\bauthor{\bsnm{Martin}, \binits{Z.K.}},
\bauthor{\bsnm{Rader}, \binits{J.K.}}:
\batitle{This {{Looks Like That There}}: {{Interpretable Neural Networks}} for
  {{Image Tasks When Location Matters}}}.
\bjtitle{Artificial Intelligence for the {E}arth Systems}
\bvolume{1}(\bissue{3}),
\bfpage{220001}
(\byear{2022}).
\doiurl{10.1175/AIES-D-22-0001.1}
\end{barticle}
\endbibitem

\bibitem{mamalakisInvestigatingFidelity2022}
\begin{barticle}
\bauthor{\bsnm{Mamalakis}, \binits{A.}},
\bauthor{\bsnm{Barnes}, \binits{E.A.}},
\bauthor{\bsnm{{Ebert-Uphoff}}, \binits{I.}}:
\batitle{Investigating the {{Fidelity}} of {{Explainable Artificial
  Intelligence Methods}} for {{Applications}} of {{Convolutional Neural
  Networks}} in {{Geoscience}}}.
\bjtitle{Artificial Intelligence for the {E}arth Systems}
\bvolume{1}(\bissue{4}),
\bfpage{220012}
(\byear{2022}).
\doiurl{10.1175/AIES-D-22-0012.1}
\end{barticle}
\endbibitem

\bibitem{pearl2017causality}
\begin{bbook}
\bauthor{\bsnm{Pearl}, \binits{J.}}:
\bbtitle{Causality: Models, Reasoning, and Inference},
\bedition{2nd} edn.
\bpublisher{MIT Press},
\blocation{Cambridge, MA}
(\byear{2017})
\end{bbook}
\endbibitem

\bibitem{Peters2017elements}
\begin{bbook}
\bauthor{\bsnm{Peters}, \binits{J.}},
\bauthor{\bsnm{Janzing}, \binits{D.}},
\bauthor{\bsnm{Sch\"olkopf}, \binits{B.}}:
\bbtitle{Elements of Causal Inference - Foundations and Learning Algorithms}.
\bsertitle{Adaptive Computation and Machine Learning Series}.
\bpublisher{MIT Press},
\blocation{Cambridge, MA}
(\byear{2017})
\end{bbook}
\endbibitem

\bibitem{Runge2023}
\begin{barticle}
\bauthor{\bsnm{Runge}, \binits{J.}},
\bauthor{\bsnm{Gerhardus}, \binits{A.}},
\bauthor{\bsnm{Varando}, \binits{G.}},
\bauthor{\bsnm{Eyring}, \binits{V.}},
\bauthor{\bsnm{Camps-Valls}, \binits{G.}}:
\batitle{{Causal inference for time series}}.
\bjtitle{Nature Reviews {E}arth and Environment}
\bvolume{4}(\bissue{7}),
\bfpage{487}--\blpage{505}
(\byear{2023}).
\doiurl{10.1038/s43017-023-00431-y}
\end{barticle}
\endbibitem

\bibitem{CampsValls23physcausaldiscovery}
\begin{barticle}
\bauthor{\bsnm{Camps-Valls}, \binits{G.}},
\bauthor{\bsnm{Gerhardus}, \binits{A.}},
\bauthor{\bsnm{Ninad}, \binits{U.}},
\bauthor{\bsnm{Varando}, \binits{G.}},
\bauthor{\bsnm{Martius}, \binits{G.}},
\bauthor{\bsnm{Balaguer-Ballester}, \binits{E.}},
\bauthor{\bsnm{Vinuesa}, \binits{R.}},
\bauthor{\bsnm{Diaz}, \binits{E.}},
\bauthor{\bsnm{Zanna}, \binits{L.}},
\bauthor{\bsnm{Runge}, \binits{J.}}:
\batitle{Discovering causal relations and equations from data}.
\bjtitle{Physics Reports}
\bvolume{1044},
\bfpage{1}--\blpage{68}
(\byear{2023})
\end{barticle}
\endbibitem

\bibitem{Gnecco2019CausalDI}
\begin{botherref}
\oauthor{\bsnm{Gnecco}, \binits{N.}},
\oauthor{\bsnm{Meinshausen}, \binits{N.}},
\oauthor{\bsnm{Peters}, \binits{J.}},
\oauthor{\bsnm{Engelke}, \binits{S.}}:
Causal discovery in heavy-tailed models.
The Annals of Statistics
(2019)
\end{botherref}
\endbibitem

\bibitem{Pasche2021CausalMO}
\begin{barticle}
\bauthor{\bsnm{Pasche}, \binits{O.C.}},
\bauthor{\bsnm{Chavez-Demoulin}, \binits{V.}},
\bauthor{\bsnm{Davison}, \binits{A.C.}}:
\batitle{Causal modelling of heavy-tailed variables and confounders with
  application to river flow}.
\bjtitle{Extremes}
\bvolume{26},
\bfpage{573}--\blpage{594}
(\byear{2021})
\end{barticle}
\endbibitem

\bibitem{Kiriliouk2019ClimateEE}
\begin{barticle}
\bauthor{\bsnm{Kiriliouk}, \binits{A.}},
\bauthor{\bsnm{Naveau}, \binits{P.}}:
\batitle{{Climate extreme event attribution using multivariate
  peaks-over-thresholds modeling and counterfactual theory}}.
\bjtitle{The Annals of Applied Statistics}
\bvolume{14}(\bissue{3}),
\bfpage{1342}--\blpage{1358}
(\byear{2020}).
\doiurl{10.1214/20-AOAS1355}
\end{barticle}
\endbibitem

\bibitem{ghanem2017}
\begin{bbook}
\bauthor{\bsnm{Ghanem}, \binits{R.}},
\bauthor{\bsnm{Higdon}, \binits{D.}},
\bauthor{\bsnm{Owhadi}, \binits{H.}}:
\bbtitle{Handbook of {Uncertainty} {Quantification}}.
\bpublisher{Springer},
\blocation{Cham, Switzerland}
(\byear{2017})
\end{bbook}
\endbibitem

\bibitem{xu2022quantifying}
\begin{barticle}
\bauthor{\bsnm{Xu}, \binits{L.}},
\bauthor{\bsnm{Chen}, \binits{N.}},
\bauthor{\bsnm{Yang}, \binits{C.}},
\bauthor{\bsnm{Yu}, \binits{H.}},
\bauthor{\bsnm{Chen}, \binits{Z.}}:
\batitle{{Quantifying the uncertainty of precipitation forecasting using
  probabilistic deep learning}}.
\bjtitle{Hydrology and {E}arth System Sciences}
\bvolume{26}(\bissue{11}),
\bfpage{2923}--\blpage{2938}
(\byear{2022}).
\doiurl{10.5194/hess-26-2923-2022}
\end{barticle}
\endbibitem

\bibitem{bella2010calibration}
\begin{bchapter}
\bauthor{\bsnm{Bella}, \binits{A.}},
\bauthor{\bsnm{Ferri}, \binits{C.}},
\bauthor{\bsnm{Hern{\'a}ndez-Orallo}, \binits{J.}},
\bauthor{\bsnm{Ram{\'\i}rez-Quintana}, \binits{M.J.}}:
\bctitle{Calibration of machine learning models}.
In: \bbtitle{Handbook of Research on Machine Learning Applications and Trends:
  Algorithms, Methods, and Techniques},
pp. \bfpage{128}--\blpage{146}.
\bpublisher{IGI Global},
\blocation{Hershey, PA}
(\byear{2010})
\end{bchapter}
\endbibitem

\bibitem{corps2023libya}
\begin{botherref}
\oauthor{\bsnm{Corps}, \binits{I.M.}}:
Libya Flooding: Situation Report \#9
(2023)
\end{botherref}
\endbibitem

\bibitem{dwd2021weather}
\begin{botherref}
\oauthor{\bsnm{DWD}}:
The weather in Germany 2021
(2021).
\url{https://www.dwd.de/EN/press/press_release/EN/2021/20210830_the_weather_in_germany_in_summer_2021_news.html}
\end{botherref}
\endbibitem

\bibitem{tojcic2021performance}
\begin{barticle}
\bauthor{\bsnm{Tojcic}, \binits{I.}},
\bauthor{\bsnm{Denamiel}, \binits{C.}},
\bauthor{\bsnm{Vilibic}, \binits{I.}}:
\batitle{{Performance of the Adriatic early warning system during the
  multi-meteotsunami event of 11–19 May 2020: an assessment using energy
  banners}}.
\bjtitle{Natural Hazards and {E}arth System Sciences}
(\byear{2021}).
\doiurl{10.5194/nhess-21-2427-2021}
\end{barticle}
\endbibitem

\bibitem{yore2020early}
\begin{barticle}
\bauthor{\bsnm{Yore}, \binits{R.}},
\bauthor{\bsnm{Walker}, \binits{J.}}:
\batitle{{Early Warning Systems and Evacuation: Rare and Extreme vs Frequent
  and Small-Scale Tropical Cyclones in the Philippines and Dominica}}.
\bjtitle{Disasters}
(\byear{2020}).
\doiurl{10.1111/disa.12434}
\end{barticle}
\endbibitem

\bibitem{reichstein2023ews}
\begin{botherref}
\oauthor{\bsnm{Reichstein}, \binits{M.}},
\oauthor{\bsnm{Benson}, \binits{V.}},
\oauthor{\bsnm{Camps-Valls}, \binits{G.}},
\oauthor{\bsnm{Boran}, \binits{H.}},
\oauthor{\bsnm{Fearnley}, \binits{C.}},
\oauthor{\bsnm{Kornhuber}, \binits{K.}},
\oauthor{\bsnm{Rahaman}, \binits{N.}},
\oauthor{\bsnm{Sch\"olkopf}, \binits{B.}},
\oauthor{\bsnm{T\'arraga}, \binits{J.M.}},
\oauthor{\bsnm{Vinuesa}, \binits{R.}}, et al.:
Early warning of complex climate risk with integrated artificial intelligence.
Nature Communications (under evaluation)
(2024)
\end{botherref}
\endbibitem

\bibitem{tamamadin2020automation}
\begin{barticle}
\bauthor{\bsnm{Tamamadin}, \binits{M.}},
\bauthor{\bsnm{Susandi}, \binits{A.}},
\bauthor{\bsnm{Pratama}, \binits{A.}},
\bauthor{\bsnm{Faisal}, \binits{I.}},
\bauthor{\bsnm{Wijaya}, \binits{A.}},
\bauthor{\bsnm{Firdaus}, \binits{I.M.}},
\bauthor{\bsnm{Kuntoro}, \binits{W.S.}}:
\batitle{Automation process to support an information system on extreme weather
  warning}.
\bjtitle{IOP Conference Series: Materials Science and Engineering}
(\byear{2020}).
\doiurl{10.1088/1757-899X/803/1/012044}
\end{barticle}
\endbibitem

\bibitem{RNY004}
\begin{botherref}
\oauthor{\bsnm{AI}, \binits{H.}}:
Ethics Guidlines for trustworthy AI: High-level expert group on artificial
  intelligence.
European Commission
(2018).
\url{https://wayback.archive-it.org/12090/20201227221227/https://ec.europa.eu/digital-single-market/en/news/ethics-guidelines-trustworthy-ai}
\end{botherref}
\endbibitem

\bibitem{RNY003}
\begin{barticle}
\bauthor{\bsnm{Jobin}, \binits{A.}},
\bauthor{\bsnm{Ienca}, \binits{M.}},
\bauthor{\bsnm{Vayena}, \binits{E.}}:
\batitle{{The global landscape of AI ethics guidelines}}.
\bjtitle{Nature Machine Intelligence}
\bvolume{1},
\bfpage{389}--\blpage{399}
(\byear{2019})
\end{barticle}
\endbibitem

\bibitem{kochupillai2022earth}
\begin{barticle}
\bauthor{\bsnm{Kochupillai}, \binits{M.}},
\bauthor{\bsnm{Kahl}, \binits{M.}},
\bauthor{\bsnm{Schmitt}, \binits{M.}},
\bauthor{\bsnm{Taubenb{\"o}ck}, \binits{H.}},
\bauthor{\bsnm{Zhu}, \binits{X.X.}}:
\batitle{Earth observation and artificial intelligence: Understanding emerging
  ethical issues and opportunities}.
\bjtitle{IEEE Geoscience and Remote Sensing Magazine}
\bvolume{10}(\bissue{4}),
\bfpage{90}--\blpage{124}
(\byear{2022})
\end{barticle}
\endbibitem

\bibitem{macherera2016review}
\begin{botherref}
\oauthor{\bsnm{Macherera}, \binits{M.}},
\oauthor{\bsnm{Chimbari}, \binits{M.J.}}:
A review of studies on community based early warning systems.
{Journal of Disaster Risk Studies}
\textbf{8}
(2016)
\end{botherref}
\endbibitem

\bibitem{Ramos2013}
\begin{barticle}
\bauthor{\bsnm{Ramos}, \binits{M.H.}},
\bauthor{\bparticle{van} \bsnm{Andel}, \binits{S.J.}},
\bauthor{\bsnm{Pappenberger}, \binits{F.}}:
\batitle{Do probabilistic forecasts lead to better decisions?}
\bjtitle{Hydrology and {E}arth System Sciences}
\bvolume{17}(\bissue{6}),
\bfpage{2219}--\blpage{2232}
(\byear{2013}).
\doiurl{10.5194/hess-17-2219-2013}
\end{barticle}
\endbibitem

\bibitem{Giuliani2015ISA}
\begin{barticle}
\bauthor{\bsnm{Giuliani}, \binits{M.}},
\bauthor{\bsnm{Pianosi}, \binits{F.}},
\bauthor{\bsnm{Castelletti}, \binits{A.}}:
\batitle{{Making the most of data: an information selection and assessment
  framework to improve water systems operations}}.
\bjtitle{Water Resources Research}
\bvolume{51}(\bissue{11}),
\bfpage{9073}--\blpage{9093}
(\byear{2015}).
\doiurl{10.1002/2015WR017044}
\end{barticle}
\endbibitem

\bibitem{camps2021deep}
\begin{bbook}
\bauthor{\bsnm{Camps-Valls}, \binits{G.}},
\bauthor{\bsnm{Tuia}, \binits{D.}},
\bauthor{\bsnm{Zhu}, \binits{X.X.}},
\bauthor{\bsnm{Reichstein}, \binits{M.}}:
\bbtitle{Deep Learning for the {E}arth Sciences: A Comprehensive Approach to
  Remote Sensing, Climate Science and Geosciences}.
\bpublisher{John Wiley \& Sons},
\blocation{Hoboken, NJ}
(\byear{2021})
\end{bbook}
\endbibitem

\bibitem{Reichstein19nat}
\begin{barticle}
\bauthor{\bsnm{Reichstein}, \binits{M.}},
\bauthor{\bsnm{Camps-Valls}, \binits{G.}},
\bauthor{\bsnm{Stevens}, \binits{B.}},
\bauthor{\bsnm{Jung}, \binits{M.}},
\bauthor{\bsnm{Denzler}, \binits{J.}},
\bauthor{\bsnm{Carvalhais}, \binits{N.}},
\bauthor{\bsnm{Prabhat}}:
\batitle{Deep learning and process understanding for data-driven {E}arth system
  science}.
\bjtitle{Nature}
\bvolume{566},
\bfpage{195}--\blpage{204}
(\byear{2019})
\end{barticle}
\endbibitem

\bibitem{iles2020benefits}
\begin{barticle}
\bauthor{\bsnm{Iles}, \binits{C.E.}},
\bauthor{\bsnm{Vautard}, \binits{R.}},
\bauthor{\bsnm{Strachan}, \binits{J.}},
\bauthor{\bsnm{Joussaume}, \binits{S.}},
\bauthor{\bsnm{Eggen}, \binits{B.R.}},
\bauthor{\bsnm{Hewitt}, \binits{C.D.}}:
\batitle{The benefits of increasing resolution in global and regional climate
  simulations for european climate extremes}.
\bjtitle{Geoscientific Model Development}
\bvolume{13}(\bissue{11}),
\bfpage{5583}--\blpage{5607}
(\byear{2020})
\end{barticle}
\endbibitem

\bibitem{cheng2023machine}
\begin{barticle}
\bauthor{\bsnm{Cheng}, \binits{S.}},
\bauthor{\bsnm{Quilodr{\'a}n-Casas}, \binits{C.}},
\bauthor{\bsnm{Ouala}, \binits{S.}},
\bauthor{\bsnm{Farchi}, \binits{A.}},
\bauthor{\bsnm{Liu}, \binits{C.}},
\bauthor{\bsnm{Tandeo}, \binits{P.}},
\bauthor{\bsnm{Fablet}, \binits{R.}},
\bauthor{\bsnm{Lucor}, \binits{D.}},
\bauthor{\bsnm{Iooss}, \binits{B.}},
\bauthor{\bsnm{Brajard}, \binits{J.}}, \betal:
\batitle{Machine learning with data assimilation and uncertainty quantification
  for dynamical systems: a review}.
\bjtitle{IEEE/CAA Journal of Automatica Sinica}
\bvolume{10}(\bissue{6}),
\bfpage{1361}--\blpage{1387}
(\byear{2023})
\end{barticle}
\endbibitem

\bibitem{Haynes2023}
\begin{botherref}
\oauthor{\bsnm{Haynes}, \binits{K.}},
\oauthor{\bsnm{Lagerquist}, \binits{R.}},
\oauthor{\bsnm{McGraw}, \binits{M.}},
\oauthor{\bsnm{Musgrave}, \binits{K.}},
\oauthor{\bsnm{Ebert-Uphoff}, \binits{I.}}:
Creating and evaluating uncertainty estimates with neural networks for
  environmental-science applications.
Artificial Intelligence for the {E}arth Systems
\textbf{2}(2)
(2023).
\doiurl{10.1175/aies-d-22-0061.1}
\end{botherref}
\endbibitem

\bibitem{Li2022}
\begin{barticle}
\bauthor{\bsnm{Li}, \binits{W.}},
\bauthor{\bsnm{Pan}, \binits{B.}},
\bauthor{\bsnm{Xia}, \binits{J.}},
\bauthor{\bsnm{Duan}, \binits{Q.}}:
\batitle{Convolutional neural network-based statistical post-processing of
  ensemble precipitation forecasts}.
\bjtitle{Journal of Hydrology}
\bvolume{605},
\bfpage{127301}
(\byear{2022}).
\doiurl{10.1016/j.jhydrol.2021.127301}
\end{barticle}
\endbibitem

\bibitem{farazmand2019extreme}
\begin{barticle}
\bauthor{\bsnm{Farazmand}, \binits{M.}},
\bauthor{\bsnm{Sapsis}, \binits{T.P.}}:
\batitle{Extreme events: Mechanisms and prediction}.
\bjtitle{Applied Mechanics Reviews}
\bvolume{71}(\bissue{5}),
\bfpage{050801}
(\byear{2019})
\end{barticle}
\endbibitem

\bibitem{west2019remote}
\begin{barticle}
\bauthor{\bsnm{West}, \binits{H.}},
\bauthor{\bsnm{Quinn}, \binits{N.}},
\bauthor{\bsnm{Horswell}, \binits{M.}}:
\batitle{Remote sensing for drought monitoring \& impact assessment: Progress,
  past challenges and future opportunities}.
\bjtitle{Remote Sensing of Environment}
\bvolume{232},
\bfpage{111291}
(\byear{2019})
\end{barticle}
\endbibitem

\bibitem{zargar2011review}
\begin{barticle}
\bauthor{\bsnm{Zargar}, \binits{A.}},
\bauthor{\bsnm{Sadiq}, \binits{R.}},
\bauthor{\bsnm{Naser}, \binits{B.}},
\bauthor{\bsnm{Khan}, \binits{F.I.}}:
\batitle{A review of drought indices}.
\bjtitle{Environmental Reviews}
\bvolume{19}(\bissue{NA}),
\bfpage{333}--\blpage{349}
(\byear{2011})
\end{barticle}
\endbibitem

\bibitem{jehanzaib2021comprehensive}
\begin{barticle}
\bauthor{\bsnm{Jehanzaib}, \binits{M.}},
\bauthor{\bsnm{Bilal~Idrees}, \binits{M.}},
\bauthor{\bsnm{Kim}, \binits{D.}},
\bauthor{\bsnm{Kim}, \binits{T.-W.}}:
\batitle{Comprehensive evaluation of machine learning techniques for
  hydrological drought forecasting}.
\bjtitle{Journal of Irrigation and Drainage Engineering}
\bvolume{147}(\bissue{7}),
\bfpage{04021022}
(\byear{2021})
\end{barticle}
\endbibitem

\bibitem{2024mengxue}
\begin{botherref}
\oauthor{\bsnm{Zhang}, \binits{M.}},
\oauthor{\bsnm{Fern\'andez-Torres}, \binits{M.A.}},
\oauthor{\bsnm{Camps-Valls}, \binits{G.}}:
Domain knowledge-driven variational recurrent networks for drought monitoring.
Remote Sensing of Environment
(2024, under review)
\end{botherref}
\endbibitem

\bibitem{dikshit2021interpretable}
\begin{barticle}
\bauthor{\bsnm{Dikshit}, \binits{A.}},
\bauthor{\bsnm{Pradhan}, \binits{B.}}:
\batitle{{Interpretable and explainable AI (XAI) model for spatial drought
  prediction}}.
\bjtitle{Science of the Total Environment}
\bvolume{801},
\bfpage{149797}
(\byear{2021})
\end{barticle}
\endbibitem

\bibitem{barriopedro_heat_2023}
\begin{botherref}
\oauthor{\bsnm{Barriopedro}, \binits{D.}},
\oauthor{\bsnm{Garc\'ia-Herrera}, \binits{R.}},
\oauthor{\bsnm{Ord\'oñez}, \binits{C.}},
\oauthor{\bsnm{Miralles}, \binits{D.}},
\oauthor{\bsnm{Salcedo-Sanz}, \binits{S.}}:
Heat waves: Physical understanding and scientific challenges.
Reviews of Geophysics
\textbf{61}(2),
2022--000780
\end{botherref}
\endbibitem

\bibitem{teng2022warming}
\begin{barticle}
\bauthor{\bsnm{Teng}, \binits{H.}},
\bauthor{\bsnm{Leung}, \binits{R.}},
\bauthor{\bsnm{Branstator}, \binits{G.}},
\bauthor{\bsnm{Lu}, \binits{J.}},
\bauthor{\bsnm{Ding}, \binits{Q.}}:
\batitle{Warming pattern over the northern hemisphere midlatitudes in boreal
  summer 1979--2020}.
\bjtitle{Journal of Climate}
\bvolume{35}(\bissue{11}),
\bfpage{3479}--\blpage{3494}
(\byear{2022})
\end{barticle}
\endbibitem

\bibitem{vautard_heat_2023}
\begin{botherref}
\oauthor{\bsnm{Vautard}, \binits{R.}},
\oauthor{\bsnm{Cattiaux}, \binits{J.}},
\oauthor{\bsnm{Happ\'e}, \binits{T.}},
\oauthor{\bsnm{Singh}, \binits{J.}},
\oauthor{\bsnm{Bonnet}, \binits{R.}},
\oauthor{\bsnm{Cassou}, \binits{C.}},
\oauthor{\bsnm{Coumou}, \binits{D.}},
\oauthor{\bsnm{D’Andrea}, \binits{F.}},
\oauthor{\bsnm{Faranda}, \binits{D.}},
\oauthor{\bsnm{Fischer}, \binits{E.}},
\oauthor{\bsnm{Ribes}, \binits{A.}},
\oauthor{\bsnm{Sippel}, \binits{S.}},
\oauthor{\bsnm{Yiou}, \binits{P.}}:
Heat extremes in western europe increasing faster than simulated due to
  atmospheric circulation trends.
Nature Communications
\textbf{14}(1),
6803.
\doiurl{10.1038/s41467-023-42143-3}.
Nature Publishing Group.
Accessed 2024-04-30
\end{botherref}
\endbibitem

\bibitem{jacques2022deep}
\begin{botherref}
\oauthor{\bsnm{Jacques-Dumas}, \binits{V.}},
\oauthor{\bsnm{Ragone}, \binits{F.}},
\oauthor{\bsnm{Borgnat}, \binits{P.}},
\oauthor{\bsnm{Abry}, \binits{P.}},
\oauthor{\bsnm{Bouchet}, \binits{F.}}:
Deep learning-based extreme heatwave forecast.
Frontiers in Climate
\textbf{4}
(2022)
\end{botherref}
\endbibitem

\bibitem{fister_accurate_2023}
\begin{botherref}
\oauthor{\bsnm{Fister}, \binits{D.}},
\oauthor{\bsnm{P\'erez-Aracil}, \binits{J.}},
\oauthor{\bsnm{Pel\'aez-Rodr\'iguez}, \binits{C.}},
\oauthor{\bsnm{Del~Ser}, \binits{J.}},
\oauthor{\bsnm{Salcedo-Sanz}, \binits{S.}}:
Accurate long-term air temperature prediction with machine learning models and
  data reduction techniques.
Applied Soft Computing
\textbf{136},
110118.
\doiurl{10.1016/j.asoc.2023.110118}.
Elsevier
\end{botherref}
\endbibitem

\bibitem{van_straaten_CorrectingSubseasonal_2023}
\begin{barticle}
\bauthor{\bparticle{van} \bsnm{Straaten}, \binits{C.}},
\bauthor{\bsnm{Whan}, \binits{K.}},
\bauthor{\bsnm{Coumou}, \binits{D.}},
\bauthor{\bparticle{van~den} \bsnm{Hurk}, \binits{B.}},
\bauthor{\bsnm{Schmeits}, \binits{M.}}:
\batitle{Correcting subseasonal forecast errors with an explainable ann to
  understand misrepresented sources of predictability of european summer
  temperatures}.
\bjtitle{Artificial Intelligence for the {E}arth Systems}
\bvolume{2}(\bissue{3}),
\bfpage{220047}
(\byear{2023}).
\doiurl{10.1175/AIES-D-22-0047.1}
\end{barticle}
\endbibitem

\bibitem{gao2023earthformer}
\begin{barticle}
\bauthor{\bsnm{Gao}, \binits{Z.}},
\bauthor{\bsnm{Shi}, \binits{X.}},
\bauthor{\bsnm{Wang}, \binits{H.}},
\bauthor{\bsnm{Zhu}, \binits{Y.}},
\bauthor{\bsnm{Wang}, \binits{Y.B.}},
\bauthor{\bsnm{Li}, \binits{M.}},
\bauthor{\bsnm{Yeung}, \binits{D.-Y.}}:
\batitle{Earthformer: Exploring space-time transformers for {E}arth system
  forecasting}.
\bjtitle{Advances in Neural Information Processing Systems}
\bvolume{35},
\bfpage{25390}--\blpage{25403}
(\byear{2022})
\end{barticle}
\endbibitem

\bibitem{singh_circulation_2023}
\begin{botherref}
\oauthor{\bsnm{Singh}, \binits{J.}},
\oauthor{\bsnm{Sippel}, \binits{S.}},
\oauthor{\bsnm{Fischer}, \binits{E.M.}}:
Circulation dampened heat extremes intensification over the midwest {USA} and
  amplified over western europe.
Communications {E}arth \& Environment
\textbf{4}(1),
1--9.
Nature Publishing Group
\end{botherref}
\endbibitem

\bibitem{jones2022global}
\begin{barticle}
\bauthor{\bsnm{Jones}, \binits{M.W.}},
\bauthor{\bsnm{Abatzoglou}, \binits{J.T.}},
\bauthor{\bsnm{Veraverbeke}, \binits{S.}},
\bauthor{\bsnm{Andela}, \binits{N.}},
\bauthor{\bsnm{Lasslop}, \binits{G.}},
\bauthor{\bsnm{Forkel}, \binits{M.}},
\bauthor{\bsnm{Smith}, \binits{A.J.P.}},
\bauthor{\bsnm{Burton}, \binits{C.}},
\bauthor{\bsnm{Betts}, \binits{R.A.}},
\bauthor{\bparticle{van~der} \bsnm{Werf}, \binits{G.R.}},
\bauthor{\bsnm{Sitch}, \binits{S.}},
\bauthor{\bsnm{Canadell}, \binits{J.G.}},
\bauthor{\bsnm{Sant{\'{i}}n}, \binits{C.}},
\bauthor{\bsnm{Kolden}, \binits{C.}},
\bauthor{\bsnm{Doerr}, \binits{S.H.}},
\bauthor{\bsnm{{Le Qu{\'{e}}r{\'{e}}}}, \binits{C.}}:
\batitle{{Global and Regional Trends and Drivers of Fire Under Climate
  Change}}.
\bjtitle{Reviews of Geophysics}
\bvolume{60}(\bissue{3}),
\bfpage{2020}--\blpage{000726}
(\byear{2022}).
\doiurl{10.1029/2020RG000726}
\end{barticle}
\endbibitem

\bibitem{el_garroussi_europe_2024}
\begin{barticle}
\bauthor{\bsnm{El~Garroussi}, \binits{S.}},
\bauthor{\bsnm{Di~Giuseppe}, \binits{F.}},
\bauthor{\bsnm{Barnard}, \binits{C.}},
\bauthor{\bsnm{Wetterhall}, \binits{F.}}:
\batitle{Europe faces up to tenfold increase in extreme fires in a warming
  climate}.
\bjtitle{npj Climate and Atmospheric Science}
\bvolume{7}(\bissue{1}),
\bfpage{1}--\blpage{11}
(\byear{2024}).
\doiurl{10.1038/s41612-024-00575-8}.
\bcomment{Nature Publishing Group}
\end{barticle}
\endbibitem

\bibitem{zhang_forest_2019}
\begin{barticle}
\bauthor{\bsnm{Zhang}, \binits{G.}},
\bauthor{\bsnm{Wang}, \binits{M.}},
\bauthor{\bsnm{Liu}, \binits{K.}}:
\batitle{Forest {Fire} {Susceptibility} {Modeling} {Using} a {Convolutional}
  {Neural} {Network} for {Yunnan} {Province} of {China}}.
\bjtitle{International Journal of Disaster Risk Science}
\bvolume{10}(\bissue{3}),
\bfpage{386}--\blpage{403}
(\byear{2019}).
\doiurl{10.1007/s13753-019-00233-1}
\end{barticle}
\endbibitem

\bibitem{li2023attentionfire_v1}
\begin{barticle}
\bauthor{\bsnm{Li}, \binits{F.}},
\bauthor{\bsnm{Zhu}, \binits{Q.}},
\bauthor{\bsnm{Riley}, \binits{W.J.}},
\bauthor{\bsnm{Zhao}, \binits{L.}},
\bauthor{\bsnm{Xu}, \binits{L.}},
\bauthor{\bsnm{Yuan}, \binits{K.}},
\bauthor{\bsnm{Chen}, \binits{M.}},
\bauthor{\bsnm{Wu}, \binits{H.}},
\bauthor{\bsnm{Gui}, \binits{Z.}},
\bauthor{\bsnm{Gong}, \binits{J.}}, \betal:
\batitle{Attentionfire\_v1. 0: interpretable machine learning fire model for
  burned-area predictions over tropics}.
\bjtitle{Geoscientific Model Development}
\bvolume{16}(\bissue{3}),
\bfpage{869}--\blpage{884}
(\byear{2023})
\end{barticle}
\endbibitem

\bibitem{fromm_understanding_2022}
\begin{barticle}
\bauthor{\bsnm{Fromm}, \binits{M.}},
\bauthor{\bsnm{Servranckx}, \binits{R.}},
\bauthor{\bsnm{Stocks}, \binits{B.J.}},
\bauthor{\bsnm{Peterson}, \binits{D.A.}}:
\batitle{Understanding the critical elements of the pyrocumulonimbus storm
  sparked by high-intensity wildland fire}.
\bjtitle{Communications {E}arth \& Environment}
\bvolume{3}(\bissue{1}),
\bfpage{1}--\blpage{7}
(\byear{2022}).
\doiurl{10.1038/s43247-022-00566-8}.
\bcomment{Nature Publishing Group}
\end{barticle}
\endbibitem

\bibitem{salas2022identifying}
\begin{bchapter}
\bauthor{\bsnm{Salas-Porras}, \binits{E.D.}},
\bauthor{\bsnm{Tazi}, \binits{K.}},
\bauthor{\bsnm{Braude}, \binits{A.}},
\bauthor{\bsnm{Okoh}, \binits{D.}},
\bauthor{\bsnm{Lamb}, \binits{K.D.}},
\bauthor{\bsnm{Watson-Parris}, \binits{D.}},
\bauthor{\bsnm{Harder}, \binits{P.}},
\bauthor{\bsnm{Meinert}, \binits{N.}}:
\bctitle{{Identifying the causes of Pyrocumulonimbus (PyroCb)}}.
In: \bbtitle{2022 NeurIPS Workshop on Causal Machine Learning for Real-World
  Impact}
(\byear{2022}).
\burl{http://arxiv.org/abs/2211.08883}
\end{bchapter}
\endbibitem

\bibitem{Jonkman2005Global}
\begin{barticle}
\bauthor{\bsnm{Jonkman}, \binits{S.}}:
\batitle{Global perspectives on loss of human life caused by floods}.
\bjtitle{Natural Hazards}
\bvolume{34},
\bfpage{151}--\blpage{175}
(\byear{2005}).
\doiurl{10.1007/S11069-004-8891-3}
\end{barticle}
\endbibitem

\bibitem{cornwall2021europe}
\begin{barticle}
\bauthor{\bsnm{Cornwall}, \binits{W.}}:
\batitle{Europe's deadly floods leave scientists stunned}.
\bjtitle{Science}
\bvolume{373}(\bissue{6553}),
\bfpage{372}--\blpage{373}
(\byear{2021}).
\doiurl{10.1126/science.373.6553.372}
\end{barticle}
\endbibitem

\bibitem{mohr2023multi}
\begin{barticle}
\bauthor{\bsnm{Mohr}, \binits{S.}},
\bauthor{\bsnm{Ehret}, \binits{U.}},
\bauthor{\bsnm{Kunz}, \binits{M.}},
\bauthor{\bsnm{Ludwig}, \binits{P.}},
\bauthor{\bsnm{Caldas-Alvarez}, \binits{A.}},
\bauthor{\bsnm{Daniell}, \binits{J.E.}},
\bauthor{\bsnm{Ehmele}, \binits{F.}},
\bauthor{\bsnm{Feldmann}, \binits{H.}},
\bauthor{\bsnm{Franca}, \binits{M.J.}},
\bauthor{\bsnm{Gattke}, \binits{C.}}, \betal:
\batitle{{A multi-disciplinary analysis of the exceptional flood event of July
  2021 in central Europe--Part 1: Event description and analysis}}.
\bjtitle{Natural Hazards and {E}arth System Sciences}
\bvolume{23},
\bfpage{525}--\blpage{551}
(\byear{2023}).
\doiurl{10.5194/nhess-23-525-2023}
\end{barticle}
\endbibitem

\bibitem{kochkov2021machine}
\begin{barticle}
\bauthor{\bsnm{Kochkov}, \binits{D.}},
\bauthor{\bsnm{Smith}, \binits{J.A.}},
\bauthor{\bsnm{Alieva}, \binits{A.}},
\bauthor{\bsnm{Wang}, \binits{Q.}},
\bauthor{\bsnm{Brenner}, \binits{M.P.}},
\bauthor{\bsnm{Hoyer}, \binits{S.}}:
\batitle{Machine learning--accelerated computational fluid dynamics}.
\bjtitle{Proceedings of the National Academy of Sciences}
\bvolume{118}(\bissue{21}),
\bfpage{2101784118}
(\byear{2021}).
\doiurl{10.1073/pnas.2101784118}
\end{barticle}
\endbibitem

\bibitem{lutjens2023physicallyconsistent}
\begin{botherref}
\oauthor{\bsnm{Lutjens}, \binits{B.}},
\oauthor{\bsnm{Leshchinskiy}, \binits{B.}},
\oauthor{\bsnm{Requena-Mesa}, \binits{C.}},
\oauthor{\bsnm{Chishtie}, \binits{F.}},
\oauthor{\bsnm{D\'iaz-Rodr\'iguez}, \binits{N.}},
\oauthor{\bsnm{Boulais}, \binits{O.}},
\oauthor{\bsnm{Sankaranarayanan}, \binits{A.}},
\oauthor{\bsnm{Masson-Forsythe}, \binits{M.}},
\oauthor{\bsnm{Piña}, \binits{A.}},
\oauthor{\bsnm{Gal}, \binits{Y.}},
\oauthor{\bsnm{Raïssi}, \binits{C.}},
\oauthor{\bsnm{Lavin}, \binits{A.}},
\oauthor{\bsnm{Newman}, \binits{D.}}:
Physically-Consistent Generative Adversarial Networks for Coastal Flood
  Visualization
(2023)
\end{botherref}
\endbibitem

\end{thebibliography}


\end{document}